\documentclass[10pt,a4paper]{article}

\usepackage[utf8]{inputenc}
\usepackage[T1]{fontenc}
\usepackage{amsmath}
\usepackage{newtxtext,newtxmath}  % Times New Roman text + matching math (incl. AMS symbols)
\usepackage{titlesec}
\usepackage{algorithm}
\usepackage{algpseudocode}
\usepackage{listings}
\usepackage{graphicx}
\usepackage[table]{xcolor}
\usepackage{colortbl}
\usepackage{booktabs}
\usepackage{array}
\newcolumntype{P}[1]{>{\raggedright\arraybackslash}p{#1}}
\usepackage{geometry}
\usepackage{hyperref}
\hypersetup{colorlinks=true, linkcolor=blue!55!black, citecolor=blue!55!black, urlcolor=blue!55!black}
\usepackage{enumitem}
\usepackage{caption}
\usepackage[numbers]{natbib}
\usepackage{float}
\usepackage{placeins}
\usepackage{tikz}
\usetikzlibrary{shapes.geometric, arrows.meta, positioning, fit, backgrounds, calc, shadows.blur,
        decorations.pathreplacing}
\usepackage{pgfplots}
\pgfplotsset{compat=1.18}
\usepgfplotslibrary{groupplots}

\definecolor{inputbg}{HTML}{36434F}
\definecolor{outputbg}{HTML}{4A6B62}
\definecolor{s1col}{HTML}{5B7594}
\definecolor{s2col}{HTML}{9A7A4E}
\definecolor{s3col}{HTML}{7B6E8C}
\definecolor{s4col}{HTML}{5F7F66}
\definecolor{s5col}{HTML}{9A5F58}
\definecolor{bugcol}{HTML}{A08A4E}
\definecolor{cardbg}{HTML}{F5F5F3}
\definecolor{ontocol}{HTML}{6E6284}
\definecolor{neo4jcol}{HTML}{53786F}
\definecolor{routecol}{HTML}{5B7594}
\tikzset{
 gcard/.style={rectangle, rounded corners=2pt, align=center, font=\scriptsize,
        minimum height=0.78cm, line width=0.4pt, inner sep=3pt},
 ginput/.style ={gcard, fill=inputbg,  draw=inputbg,   text=white},
 goutput/.style={gcard, fill=outputbg, draw=outputbg,   text=white},
 gonto/.style ={gcard, fill=ontocol!12, draw=ontocol!65, text=black!80},
 groute/.style ={gcard, fill=routecol!12, draw=routecol!65, text=black!80},
 gproc/.style ={gcard, fill=black!3,  draw=s1col!60,   text=black!80},
 gproc2/.style ={gcard, fill=s3col!10, draw=s3col!60,   text=black!80},
 gdata/.style ={gcard, fill=white,   draw=black!45,   text=black!78},
 gmerge/.style ={gcard, fill=s4col!10, draw=s4col!60,   text=black!80},
 greview/.style={gcard, fill=bugcol!12, draw=bugcol!70,  text=black!80},
 gdistinct/.style={gcard, fill=black!4, draw=black!35,   text=black!65},
 gdb/.style={cylinder, shape border rotate=90, aspect=0.25, draw=neo4jcol!70,
       fill=neo4jcol!8, text=black!78, align=center, font=\scriptsize,
       line width=0.4pt, minimum width=2.1cm, minimum height=1.25cm},
 gdiamond/.style={diamond, aspect=2.2, draw=s2col!70, fill=s2col!10,
          text=black!80, align=center, font=\scriptsize, inner sep=1pt,
          line width=0.4pt},
 gnote/.style={rectangle, rounded corners=2pt, draw=black!25, fill=black!3,
        font=\scriptsize, align=center, inner sep=4pt, line width=0.4pt},
 gflow/.style={-{Stealth[length=1.9mm, width=1.4mm]}, line width=0.5pt, draw=black!55},
 grec/.style={rectangle, rounded corners=2pt, draw=black!40, fill=white,
        font=\scriptsize, inner sep=4pt, line width=0.4pt},
 grecbad/.style={grec, draw=s5col!70, fill=s5col!5},
 grecok/.style={grec, draw=s4col!70, fill=s4col!5},
 ghdr/.style={font=\scriptsize\bfseries, text=black!70},
 gtiny/.style={font=\tiny, text=black!55},
}

\makeatletter
\renewcommand\normalsize{%
 \@setfontsize\normalsize{10}{11}%
 \abovedisplayskip 9\p@ \@plus2\p@ \@minus5\p@
 \belowdisplayskip \abovedisplayskip
 \abovedisplayshortskip \z@ \@plus3\p@
 \belowdisplayshortskip 6\p@ \@plus3\p@ \@minus3\p@}
\makeatother
\normalsize
\titleformat{\section}
 {\normalfont\fontsize{12}{14}\selectfont\bfseries}{\thesection}{1em}{}
\titleformat{\subsection}
 {\normalfont\fontsize{10}{12}\selectfont\bfseries}{\thesubsection}{1em}{}
\titleformat{\subsubsection}
 {\normalfont\fontsize{10}{12}\selectfont\bfseries}{\thesubsubsection}{1em}{}
\titlespacing*{\section}{0pt}{2.0ex plus 0.5ex minus 0.2ex}{1.0ex plus 0.2ex}
\titlespacing*{\subsection}{0pt}{1.6ex plus 0.4ex minus 0.2ex}{0.8ex plus 0.2ex}
\titlespacing*{\subsubsection}{0pt}{1.3ex plus 0.3ex minus 0.2ex}{0.6ex plus 0.2ex}

\definecolor{codebg}{RGB}{245,245,245}
\definecolor{codegreen}{RGB}{0,128,0}
\definecolor{codegray}{RGB}{128,128,128}
\definecolor{codepurple}{RGB}{128,0,128}

\lstdefinestyle{pythonstyle}{
  backgroundcolor=\color{codebg},
  commentstyle=\color{codegreen},
  keywordstyle=\color{blue},
  stringstyle=\color{codepurple},
  basicstyle=\ttfamily\footnotesize,
  breaklines=true,
  frame=single,
  rulecolor=\color{codegray},
  numbers=left,
  numberstyle=\tiny\color{codegray},
  language=Python,
  showstringspaces=false,
  tabsize=4
}

\lstdefinestyle{jsonstyle}{
  backgroundcolor=\color{codebg},
  basicstyle=\ttfamily\footnotesize,
  breaklines=true,
  frame=single,
  rulecolor=\color{codegray},
  numbers=left,
  numberstyle=\tiny\color{codegray},
  showstringspaces=false,
  tabsize=2
}

\newcommand{\fitwidth}[1]{%
 \resizebox{\ifdim\width>\linewidth \linewidth\else \width\fi}{!}{#1}}

\newif\ifanon
\anonfalse

\newcommand{\titletext}{Curate Before You Connect: Identity and Ontology Tagging in a Production Knowledge Graph}

\newcommand{\makenipstitle}{%
 \par\noindent\hrule height 4pt\relax
 \vspace{0.25in}
 \begin{center}
  {\fontsize{17}{21}\selectfont\bfseries \titletext\par}
 \end{center}
 \vspace{0.25in}
 \par\noindent\hrule height 1pt\relax
 \vspace{0.30in}
 \begin{center}
  \ifanon
   {\fontsize{12}{14}\selectfont\bfseries Anonymous Author(s)}\\[2pt]
   {\fontsize{10}{12}\selectfont Affiliation\\Address\\\texttt{email}}\\[2pt]
   {\fontsize{10}{12}\selectfont Paper under double-blind review}
  \else
   \begin{tabular}{c@{\hskip 1.4em}c@{\hskip 1.4em}c}
    {\fontsize{12}{14}\selectfont\bfseries Vaibhav Dangaich} &
    {\fontsize{12}{14}\selectfont\bfseries Kevin Lewis} &
    {\fontsize{12}{14}\selectfont\bfseries Kundeshwar Pundalik} \\[3pt]
    {\fontsize{8.5}{10}\selectfont\texttt{vaibhavdangaich@gmail.com}} &
    {\fontsize{8.5}{10}\selectfont\texttt{kevin.lewis@konectu.in}} &
    {\fontsize{8.5}{10}\selectfont\texttt{kundeshwar@konectu.in}}
   \end{tabular}
  \fi
 \end{center}
 \vspace{0.25in}
}
\date{}

\begin{document}

\makenipstitle
\thispagestyle{empty}

\begin{center}
 {\bfseries\large Abstract}
\end{center}
\vspace{-0.5em}
\begin{quote}
Extraction produces candidate entities and relationships; writing them into a graph is where identity is decided, and identity decisions are destructive in a way extraction errors are not. A wrong type can be corrected later, but two records merged under one identity cannot be separated once their properties have been combined, and the merge leaves no error behind. This paper describes the ingestion and ontology-tagging layer that turns a validated extraction stream into a knowledge graph of 537{,}157 entities and 2{,}198{,}567 relationships drawn from 98{,}795 government documents.

We describe a record-identity ladder that decides sameness from identifier columns, name columns, display names and type-scoped position rather than from name similarity. The ladder governs de-duplication within parsed tables, while the graph write applies a coarser canonical-name key, so records sharing a canonical name merge automatically on exact equality. We argue rather than demonstrate that this is where the automation line belongs: no identity benchmark is reported, and the over-merges the key permits are undetectable by construction. That policy, under which entity resolution only ever flags candidates, followed an incident in which two surface forms of one name were merged, corrupting a correct record and deleting eight entities from an unrelated document.

We then describe multi-class ontology tagging and an evidence asymmetry we did not anticipate: an entity name is an instance label rather than a type assertion, so matching name fragments against a class index invents classifications. Requiring anchored evidence cut role assignments on an enriched sample from 36 to 4, all confirmed correct. We quantify the graph's conformance debt, show secondary classifications compensating for a mis-parented primary class, and describe a curation queue grown to 48{,}403 pending proposals against 775 human decisions.
\end{quote}

\vspace{0.4em}

\clearpage

% ══════════════════════════════════════════════════════════════
\section{Introduction}

A companion paper describes the extraction layer that converts a heterogeneous
document stream into validated entities and relationships. This paper begins
where that one ends: with extraction output in hand, and the problem of writing
it into a graph that will be queried \cite{hogan2021knowledge, angles2008survey}.

That step is usually treated as plumbing. We found it is where the consequential
decisions are made. Extraction errors are recoverable: a mistyped entity can
be retyped, a missed relationship can be re-extracted on a later pass. Identity
errors are not \cite{christophides2020overview, getoor2012entity}. Once two
records have been merged under a single node and their
properties combined, the information required to separate them has been
destroyed. Worse, the operation is silent: no constraint is violated, no
exception is raised, and the resulting node is perfectly well-formed. It is
simply describing something that does not exist. Faults of this shape are the
ones that travel furthest before anyone notices, because every stage downstream
treats them as data \cite{sambasivan2021cascades}.

This asymmetry drives the whole design. Where a decision is mechanical we
automate it without hesitation; where it requires judgement we escalate it; and
where an operation is both destructive and hard to detect we decline to automate
it at any confidence threshold. Section~\ref{sec:incident} describes the
incident that taught us the third rule.

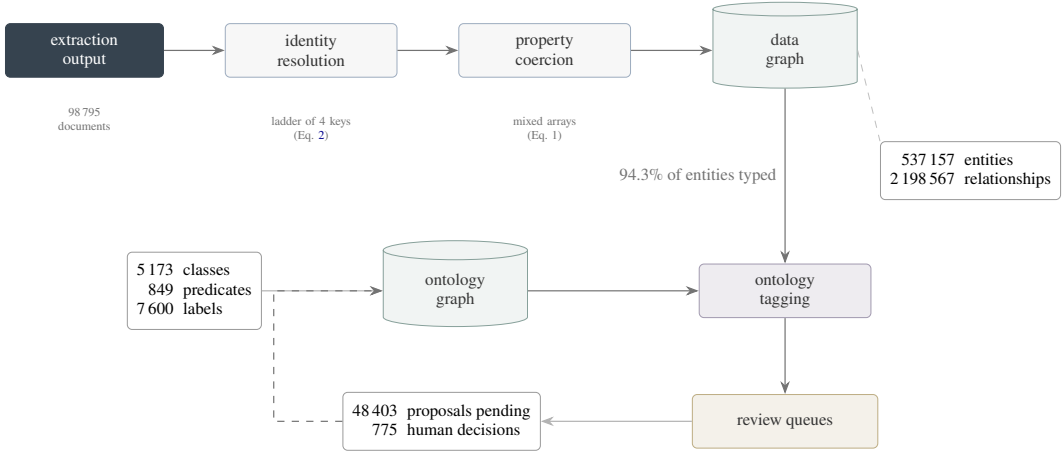
\begin{figure}[H]
\centering
\fitwidth{%
\begin{tikzpicture}
 \node[ginput, minimum width=2.3cm] (ext) at (0,0) {extraction\\output};
 \node[gtiny, align=center] at (0,-1.0) {98\,795\\documents};

 \node[gproc, minimum width=2.5cm] (ident) at (3.3,0) {identity\\resolution};
 \node[gtiny, text width=2.5cm, align=center] at (3.3,-1.15) {ladder of 4 keys\\(Eq.~\ref{eq:identity})};

 \node[gproc, minimum width=2.5cm] (coerce) at (6.7,0) {property\\coercion};
 \node[gtiny, text width=2.5cm, align=center] at (6.7,-1.15) {mixed arrays\\(Eq.~1)};

 \node[gdb] (graph) at (10.2,0) {data\\graph};
 \node[grec] (gstats) at (12.9,-1.75) {\begin{tabular}{@{}r@{\hskip 4pt}l@{}}
  537\,157 & entities\\ 2\,198\,567 & relationships\\
 \end{tabular}};

 \node[gonto, minimum width=2.5cm] (tag) at (10.2,-3.5) {ontology\\tagging};
 \node[gdb] (onto) at (5.4,-3.5) {ontology\\graph};
 \node[grec] (ostats) at (1.6,-3.5) {\begin{tabular}{@{}r@{\hskip 4pt}l@{}}
  5\,173 & classes\\ 849 & predicates\\ 7\,600 & labels\\
 \end{tabular}};

 \node[greview, minimum width=2.7cm] (queue) at (10.2,-5.4) {review queues};
 \node[grec] (qstats) at (5.2,-5.4) {\begin{tabular}{@{}r@{\hskip 4pt}l@{}}
  48\,403 & proposals pending\\ 775 & human decisions\\
 \end{tabular}};

 \draw[gflow] (ext) -- (ident);
 \draw[gflow] (ident) -- (coerce);
 \draw[gflow] (coerce) -- (graph);
 \draw[gflow] (graph) -- node[left, font=\scriptsize, text=black!55] {94.3\% of entities typed} (tag);
 \draw[gflow] (onto) -- (tag);
 \draw[draw=black!25, line width=0.4pt, dashed] (graph.east) -- (gstats.north west);
 \draw[gflow, draw=black!30] (ostats) -- (onto);
 \draw[gflow] (tag) -- (queue);
 \draw[gflow, draw=black!30] (queue) -- (qstats);
 \draw[gflow, dashed] (qstats.west) -- ++(-1.0,0) |- (onto.west);
\end{tikzpicture}}
\caption{The ingestion and tagging layer. Solid arrows carry data; the dashed
arrow is the curation loop, in which human decisions modify the ontology that
subsequently governs tagging. Retrieval over the resulting graph is the subject
of a separate paper.}
\label{fig:overview}
\end{figure}

\subsection{Contributions}

\begin{enumerate}[leftmargin=*, itemsep=2pt, topsep=2pt]
 \item A \textbf{record-identity ladder} for de-duplication within parsed
    tables, deciding sameness from identifier columns, name columns, display
    names, and type-scoped position, in that order, rather than from name
    similarity (\S\ref{sec:identity}). The graph write applies a separate
    canonical-name key, which we state and defend rather than measure: the
    paper reports no identity benchmark, and \S\ref{sec:limits} says what
    one would need to contain.
 \item An \textbf{incident analysis} of an automated merge that corrupted a
    correct record and deleted eight entities from an unrelated document,
    and the flag-only resolution policy adopted in response
    (\S\ref{sec:incident}, \S\ref{sec:flagonly}).
 \item \textbf{Multi-class ontology tagging}, in which an entity carries
    several non-subsuming classes concurrently (\S\ref{sec:multiclass}),
    together with the \textbf{name-versus-type evidence asymmetry} that
    governs when a class may be inferred from an entity's name, and a
    measurement of its effect on an enriched sample
    (\S\ref{sec:asymmetry}).
 \item A \textbf{quantification of conformance debt} in a production graph of
    2.2M relationships, including a case in which secondary classifications
    were compensating for a mis-parented primary class
    (\S\ref{sec:conformance}).
 \item The design of a \textbf{checkpointed triage agent} for the ontology
    proposal queue, in which a language model proposes and a deterministic
    validator disposes, with dry-run defaults and per-run reversal
    (\S\ref{sec:agent}).
\end{enumerate}

Because the measurements in this paper come from three different sources, we
give their provenance once here rather than repeating it in footnotes.

\begin{table}[H]
\centering
\caption{Where each measurement comes from and how far it can be trusted. No
number in this paper was taken on a quiescent graph, and none of the three
corpora exercises the whole design.}
\label{tab:provenance}
\small
\begin{tabular}{@{}P{3.5cm}P{3.0cm}P{3.0cm}P{2.6cm}@{}}
\toprule
Claim & Corpus & Status & Graph quiescent \\
\midrule
Graph scale, tagging coverage (Table~\ref{tab:coverage}) & Maharashtra government resolutions & Single run, relationship pass interrupted & No, tagger writing \\
Conformance counts (Table~\ref{tab:debt-counts}) & Maharashtra GRs & Single run; floors, occurrences not distinct terms & No \\
Queue sizes (\S\ref{sec:flagonly}, \S\ref{sec:queue}) & Maharashtra GRs & Point-in-time; window unrecorded & Not applicable \\
Evidence-rule precision (Figure~\ref{fig:evidence}) & Maharashtra GRs & Enriched sample, $n=2{,}500$, non-uniform & No \\
Identity ladder, incident, worked resolution & Defence document set with declared table schemas & Argued, not benchmarked; incident reconstructed & Not applicable \\
Edge-identity fix, insertion path (\S\ref{sec:edgeid}, \S\ref{sec:insertionpath}) & Separate smaller evaluation corpus & As measured there; pending re-measurement & Unknown \\
Triage agent (\S\ref{sec:agent}) & n/a & Designed and scaffolded; never run & Not applicable \\
\bottomrule
\end{tabular}
\end{table}

% ══════════════════════════════════════════════════════════════
\section{Ingestion into the graph}
\label{sec:ingest}

\subsection{Idempotency and re-ingest semantics}

Ingestion is driven from a spool directory of extraction outputs and tracked by
an append-only manifest. A document already named in the manifest is skipped, so
a run interrupted at any point resumes without duplicating work, and re-running
a completed batch is a no-op rather than a second insertion
\cite{helland2012idempotence}.

The manifest is never pruned, which is deliberate (it is the record of what
was ever ingested, not of what is currently on disk) but it has a consequence
worth stating because it produced a false alarm in operation. Documents that
were ingested and later moved out of the spool remain in the manifest while
disappearing from the directory listing. A progress indicator computed as
$|{\rm manifest}| / |{\rm spool}|$ therefore reported $98{,}860/98{,}795$: a
completion ratio above one, which reads as a fault when nothing is wrong. The
correct denominator is the intersection of manifest and spool, with the
remainder reported separately as archived.

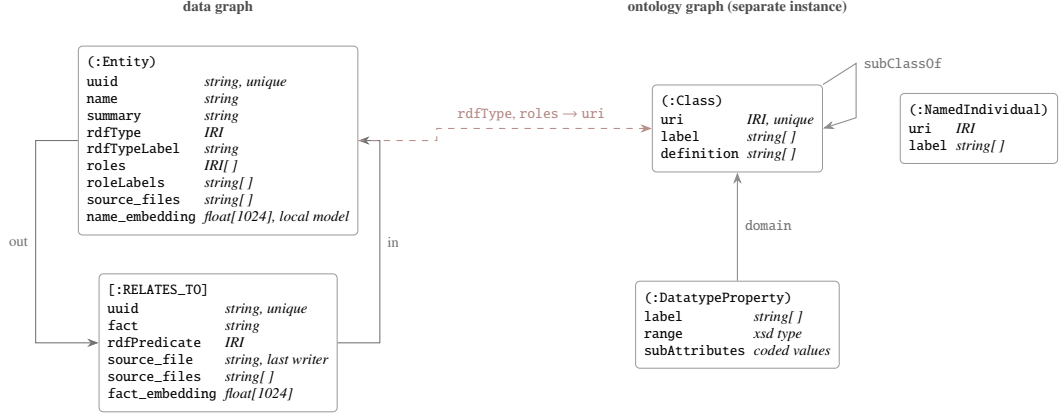
\begin{figure}[H]
\centering
\fitwidth{%
\begin{tikzpicture}
 \node[ghdr] at (0,2.95) {data graph};
 \node[grec] (ent) at (0,0.75) {\begin{tabular}{@{}l@{\hskip 4pt}l@{}}
  \multicolumn{2}{@{}l@{}}{\texttt{(:Entity)}}\\[1pt]
  \texttt{uuid} & \textit{string, unique}\\
  \texttt{name} & \textit{string}\\
  \texttt{summary} & \textit{string}\\
  \texttt{rdfType} & \textit{IRI}\\
  \texttt{rdfTypeLabel} & \textit{string}\\
  \texttt{roles} & \textit{IRI[\,]}\\
  \texttt{roleLabels} & \textit{string[\,]}\\
  \texttt{source\_files} & \textit{string[\,]}\\
  \texttt{name\_embedding} & \textit{float[1024], local model}\\
 \end{tabular}};

 \node[grec] (rel) at (0,-2.6) {\begin{tabular}{@{}l@{\hskip 4pt}l@{}}
  \multicolumn{2}{@{}l@{}}{\texttt{[:RELATES\_TO]}}\\[1pt]
  \texttt{uuid} & \textit{string, unique}\\
  \texttt{fact} & \textit{string}\\
  \texttt{rdfPredicate} & \textit{IRI}\\
  \texttt{source\_file} & \textit{string, last writer}\\
  \texttt{source\_files} & \textit{string[\,]}\\
  \texttt{fact\_embedding} & \textit{float[1024]}\\
 \end{tabular}};

 \draw[gflow] (ent.west) -- ++(-0.7,0) |- node[left, font=\scriptsize, text=black!55, pos=0.25] {out} (rel.west);
 \draw[gflow] (rel.east) -- ++(0.7,0) |- node[right, font=\scriptsize, text=black!55, pos=0.25] {in} (ent.east);

 \node[ghdr] at (8.6,2.95) {ontology graph (separate instance)};
 \node[grec] (cls) at (8.6,0.95) {\begin{tabular}{@{}l@{\hskip 4pt}l@{}}
  \multicolumn{2}{@{}l@{}}{\texttt{(:Class)}}\\[1pt]
  \texttt{uri} & \textit{IRI, unique}\\
  \texttt{label} & \textit{string[\,]}\\
  \texttt{definition} & \textit{string[\,]}\\
 \end{tabular}};
 \node[grec] (ni) at (12.6,0.95) {\begin{tabular}{@{}l@{\hskip 4pt}l@{}}
  \multicolumn{2}{@{}l@{}}{\texttt{(:NamedIndividual)}}\\[1pt]
  \texttt{uri} & \textit{IRI}\\
  \texttt{label} & \textit{string[\,]}\\
 \end{tabular}};
 \node[grec] (dp) at (8.6,-2.3) {\begin{tabular}{@{}l@{\hskip 4pt}l@{}}
  \multicolumn{2}{@{}l@{}}{\texttt{(:DatatypeProperty)}}\\[1pt]
  \texttt{label} & \textit{string[\,]}\\
  \texttt{range} & \textit{xsd type}\\
  \texttt{subAttributes} & \textit{coded values}\\
 \end{tabular}};

 \draw[gflow, draw=black!45] (cls.north east) -- ++(0.55,0.35)
    node[right, font=\scriptsize, text=black!55] {\texttt{subClassOf}} -- ++(0,-0.9) -- (cls.east);
 \draw[gflow, draw=black!45] (dp.north) -- node[right, font=\scriptsize, text=black!55] {\texttt{domain}} (cls.south);

 \draw[gflow, dashed, draw=s5col!70] (ent.east) -- ++(1.3,0) |-
    node[above, font=\scriptsize, text=s5col!75, pos=0.72] {\texttt{rdfType}, \texttt{roles} $\rightarrow$ \texttt{uri}} (cls.west);
\end{tikzpicture}}
\caption{The stored schema. The two databases are joined only by IRI reference
(dashed): the data graph records which ontology classes an entity carries, but
holds no copy of the ontology itself. Array-valued \texttt{roles} and
\texttt{roleLabels} are what permit an entity to hold several classifications
at once (\S\ref{sec:multiclass}). Nodes and edges both carry
\texttt{source\_files}: because Equation~\ref{eq:edgeid} deliberately merges
one fact asserted in several documents onto a single edge, a scalar field could
not hold its provenance, so each merge appends to the array if the document is
not already present.}
\label{fig:schema}
\end{figure}

\subsection{Property coercion}

Neo4j \cite{robinson2015graph, francis2018cypher} stores homogeneous arrays
only: a list property whose elements are of
mixed type is rejected at write time, and the rejection fails the entire node,
not the offending property. Extraction output does contain such lists: a confidence array carrying both floats and the string \texttt{"n/a"}, for
instance. The ingestion layer inspects the element kinds of every list-valued
property and, when more than one kind is present, stringifies the whole list
rather than dropping the property or failing the node:

\begin{equation}
\mathrm{coerce}(L) =
\begin{cases}
L, & |\{\,\tau(x): x \in L\,\}| \le 1 \\[2pt]
[\,\mathrm{str}(x): x \in L\,], & \text{otherwise}
\end{cases}
\end{equation}

where $\tau$ maps a value to one of $\{\textsf{bool}, \textsf{str},
\textsf{num}\}$. Booleans are classified before numbers, since in Python
\texttt{bool} is a subclass of \texttt{int} and the natural test silently
merges the two kinds: an ordering bug we shipped before finding it.

\subsection{The record-identity ladder}
\label{sec:identity}

Before a record can be written it must be matched against what is already in the
graph \cite{christen2012matching}. Matching on name alone is the obvious
approach, and it fails in both
directions. Two records may name the same entity differently, and (less
obviously) two genuinely distinct records may carry identical names, because
the extractor hallucinated one from the other or because the source document
reuses a label across rows.

We therefore resolve identity through an ordered ladder of keys
\cite{christophides2020overview, fellegi1969theory, elmagarmid2007duplicate},
taking the
first that is available. For a record $r$ of entity type $t$:

\begin{equation}
\kappa(r) =
\begin{cases}
(\textsf{id}, r.\mathrm{id}), & \text{if an identifier column is declared and populated} \\
(\textsf{name}, \mathrm{lower}(r.\mathrm{name}_{\mathrm{col}})), & \text{else if a name column is declared} \\
(\textsf{disp}, \mathrm{lower}(r.\mathrm{display})), & \text{else if a display name exists} \\
(\textsf{geo}\!:\!t,\; \mathrm{lat},\mathrm{lon}), & \text{else if the record is positioned} \\
\bot, & \text{otherwise}
\end{cases}
\label{eq:identity}
\end{equation}

Scope matters here, and we have been imprecise about it. The identifier and
name rungs are not free-text fields: they are columns declared in a table
schema, so an identifier is a domain identifier of a known record type (\texttt{TRACK\_ID}, \texttt{ALERT\_ID}, \texttt{NODE\_ID}) and a name
column is that schema's descriptive column. The ladder fires where a parser has
recognised a table and knows which of its columns identify a row. It is not a
general cross-corpus merge of anything that shares a label.

A second key applies at the graph write, and it is not the same key. Node
identity is computed from the canonicalised name alone, scoped by type only when
that name is generic (\emph{Headquarters}, \emph{Facility}) where an
unscoped key would collapse unrelated things. Distinctive names are keyed by
name without the type.

Two consequences follow, and we state them because they are not what
Equation~\ref{eq:identity} on its own would lead a reader to expect. First, the
identifier rung's separating power does not survive into the graph: two records
carrying distinct declared identifiers but the same canonical name are matched
within the parsed table, and then written as one node. The ladder governs
de-duplication inside extraction; the graph write re-keys on the name. Second,
records sharing a canonical name therefore merge automatically on exact
equality, so in a corpus where \emph{Collector} or \emph{Under Secretary}
recurs across thousands of documents those records become one office: usually
what a reader wants, and wrong when two individuals share a name.

The criterion we are applying is worth stating positively rather than as an
apology. We automate a merge when its warrant is checkable and reproducible (exact equality on a canonicalised string) and refuse when the warrant is a
score over free text, because a score cannot be audited after the fact and a
string comparison can. That is the line, and canonical-name merging sits on the
automated side of it by that test.

What we should not claim is that anything catches the residue. Two officials
sharing a name and a type are merged before entity resolution sees them, so
there is no candidate pair left to rank; the queue of \S\ref{sec:flagonly}
surfaces the opposite failure, two nodes that should be one, and its $1{,}002$
pairs are the cost of \emph{not} merging rather than of merging. Nothing in the
system currently detects an over-merge of this kind. We think the case is rare
in a corpus dominated by offices rather than persons, but that is a judgement
about the corpus, not a mechanism.

Two properties of Equation~\ref{eq:identity} are load-bearing. First, the ladder
is ordered by \emph{authority}, not by availability: a declared identifier
column beats a matching name, so two rows that share a name but carry different
identifiers stay separate. Second, the positional key is scoped by entity type.
Position alone is not identity (a sensor and the facility it sits inside
share coordinates without being the same thing) so the type is part of the
key rather than a filter applied afterwards.

The positional key compares coordinates as written, with no tolerance. That is
deliberate but narrow: it catches verbatim reproduction, which is the failure it
was built for, and it will not match two records giving one place to different
precision. A tolerance needs a policy for how near is near enough, which between
adjacent facilities is a judgement rather than a threshold, so we did not
introduce one.

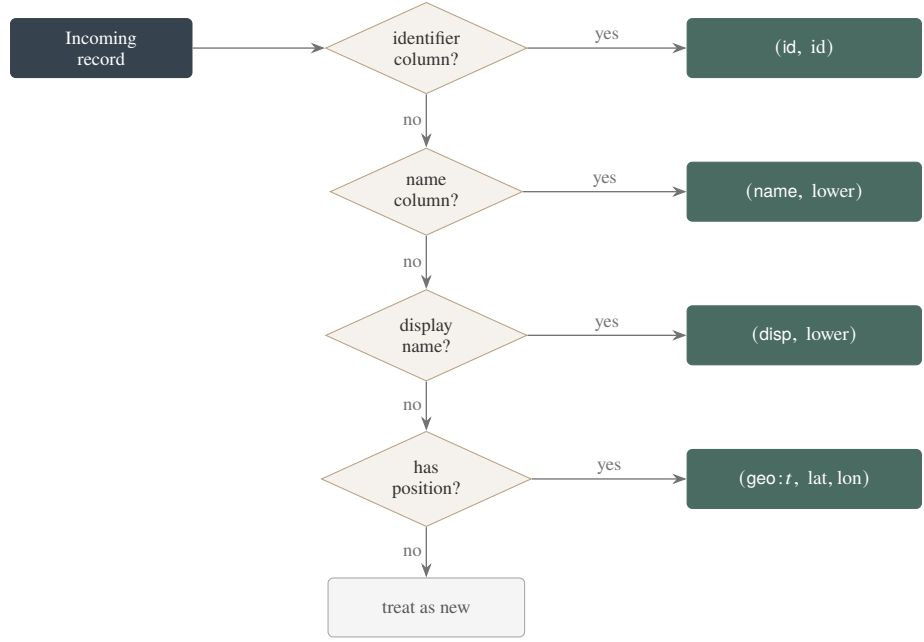
\begin{figure}[H]
\centering
\fitwidth{%
\begin{tikzpicture}
 \node[ginput, minimum width=2.4cm] (rec) at (0,0) {Incoming\\record};
 \foreach \i/\y/\q in {1/0/{identifier\\column?}, 2/-1.9/{name\\column?},
            3/-3.8/{display\\name?}, 4/-5.7/{has\\position?}}
  \node[gdiamond, text width=1.45cm] (d\i) at (4.3,\y) {\q};
 \node[gdistinct, minimum width=2.6cm] (new) at (4.3,-7.4) {treat as new};

 \node[goutput, minimum width=3.1cm] (k1) at (9.3,0)  {$(\textsf{id},\ \mathrm{id})$};
 \node[goutput, minimum width=3.1cm] (k2) at (9.3,-1.9) {$(\textsf{name},\ \mathrm{lower})$};
 \node[goutput, minimum width=3.1cm] (k3) at (9.3,-3.8) {$(\textsf{disp},\ \mathrm{lower})$};
 \node[goutput, minimum width=3.1cm] (k4) at (9.3,-5.7) {$(\textsf{geo}\!:\!t,\ \mathrm{lat},\mathrm{lon})$};

 \draw[gflow] (rec) -- (d1);
 \foreach \i in {1,2,3,4}
  \draw[gflow] (d\i) -- node[above, font=\scriptsize, text=black!55, inner sep=1.5pt] {yes} (k\i);
 \foreach \a/\b in {d1/d2, d2/d3, d3/d4, d4/new}
  \draw[gflow] (\a) -- node[left, font=\scriptsize, text=black!55, inner sep=1.5pt] {no} (\b);
\end{tikzpicture}}
\caption{The record-identity ladder. The first available key wins; authority
decreases downward. The positional key carries the entity type $t$ so that two
different kinds of thing at one location do not collide.}
\label{fig:ladder}
\end{figure}

The positional rung earns its place on rows that arrive without an identifier
and without a matching name, which the extraction leg produces regularly: a row
carrying coordinates and nothing else that any earlier rung can match.
\S\ref{sec:limits} records the case that motivated it, which is better read as
a limitation of the ordering than as a rationale for the rung.

\begin{algorithm}[H]
\caption{Ingest a record under the identity ladder}
\label{alg:upsert}
\begin{algorithmic}
\Require record $r$ of declared type $t$; graph $G$
\State $k \gets \kappa(r)$ \Comment{Equation~\ref{eq:identity}; $\bot$ if no rung applies}
\If{$k = \bot$}
 \State \Return \Call{Insert}{$G$, $r$} \Comment{unidentifiable: never merged}
\EndIf
\State $M \gets \{\, n \in G \;:\; \mathrm{type}(n) = t \;\wedge\; \kappa(n) = k \,\}$
\If{$|M| = 0$}
 \State \Return \Call{Insert}{$G$, $r$}
\ElsIf{$|M| = 1$}
 \State $r' \gets{}$ \Call{Coerce}{$r$} \Comment{property coercion}
 \State \Return \Call{UpdateProperties}{$M_1$, $r'$}
\Else
 \State \Call{FlagForReview}{$M$, $r$} \Comment{ambiguous key: escalate, do not merge}
 \State \Return \Call{Insert}{$G$, $r$}
\EndIf
\end{algorithmic}
\end{algorithm}

\paragraph{What \textsc{UpdateProperties} does on conflict.} The update branch
combines properties too, which is the operation that caused the damage in
\S\ref{sec:incident}, so its semantics need stating. Scalar properties are
last-write-wins: a later document's value replaces an earlier one. Provenance is
the exception and accumulates, since a fact asserted by two documents is
evidenced by both. A small set of fields (name, summary and aliases) is
protected once a human has edited the node, so bulk ingestion cannot silently
overwrite a curator.

Last-write-wins is defensible here and was not defensible in
\S\ref{sec:incident}, and the difference is not the merge policy but the key.
Combining the properties of two records that a declared identifier says are the
same record loses a stale value. Combining the properties of two records that a
similarity score \emph{guessed} were the same loses a real one. The operation is
identical; only the warrant differs.

The final branch is the whole policy in one statement. An ambiguous key is the exact
situation in which a similarity matcher would have guessed, and it is the
situation in which guessing is least recoverable, so the record is written
separately and the collision is escalated.

\subsection{Edge identity}
\label{sec:edgeid}

Nodes are not the only things that need identities. An edge does too, and getting
it wrong duplicates the relationship layer rather than the entity layer: which
is harder to notice, because a duplicated edge between two correct nodes looks
entirely reasonable.

Our first implementation derived an edge's identity by hashing the relation type
together with the identifier of the \emph{file} the relationship was read from.
That is provenance, not identity. The same relationship stated in two documents
produced two edges; the same relationship stated twice within one document
produced two more. Identity now comes from the resolved endpoints and the
normalised relation type, and from nothing else:

\begin{equation}
\mathrm{id}(e) = H\bigl(\,\mathrm{uuid}(\mathrm{src}) \;\Vert\;
             \mathrm{uuid}(\mathrm{tgt}) \;\Vert\;
             \mathrm{norm}(\mathrm{type})\,\bigr)
\label{eq:edgeid}
\end{equation}

Because the endpoint identities are themselves resolved by
Equation~\ref{eq:identity}, and because none of the three inputs mentions a file,
the same fact asserted anywhere in the corpus lands on the same edge and is
merged rather than repeated. On a separately evaluated corpus the correction
reduced the edge count from 10{,}757 to 8{,}350: 2{,}407
spurious edges, or 22.4\% of the relationship layer as it stood before the fix.
Stated the other way, file-scoped identities inflated that layer by 28.8\% over
its corrected size. The two percentages describe the same 2{,}407 edges against
different denominators, and we give the pre-fix rate because that is the
proportion of stored edges that were wrong.\footnote{Figures in this subsection and in
\S\ref{sec:insertionpath} come from a separate, smaller evaluation than the
corpus of Table~\ref{tab:coverage}; they are reported as measured there and are
pending re-measurement (\S\ref{sec:limits}). The production layer of
Table~\ref{tab:coverage} does not carry the defect described here:
content-addressed edge identity was in place months before that corpus was
ingested, so its $2{,}198{,}567$ relationships, and the $51.9\%$ predicate
coverage computed over them, contain no file-scoped duplication.}

The principle generalises beyond this system: \emph{identity must be derived
from what a thing is, never from the circumstances under which it was
encountered}. File paths, object identifiers and ingestion timestamps belong in
an audit trail. As inputs to a de-duplication hash they guarantee the duplicates
they were meant to prevent.

\subsection{An identity failure and what it cost}
\label{sec:incident}

The ladder replaced a similarity-based matcher, and it did so because that
matcher caused a production incident that we describe here in full, because the
detail is the argument.

A power-grid sector appeared in two documents, once with a spaced name and once
with the same name underscored. A similarity matcher scored these as the same
entity (correctly, by any string metric \cite{cohen2003string}) and merged
them. The merge
combined properties from both records. The surviving node acquired coordinates
belonging to a different facility, a coverage radius of $95{,}000$ metres, an
entity type of \emph{Location} in place of its correct type, and an
\texttt{is\_illustrative} flag marking it as synthetic. Before the merge the
node had been correct.

The merge additionally deleted eight entities belonging to an unrelated
document. The mechanism is worth spelling out, because it is the part that
turns a bad classification into data loss. Provenance in this system is
per-document: a node records the files it was derived from, and the maintenance
path that removes a document's contribution works by walking that provenance.
When the merge combined the two records it also combined their provenance, so
the surviving node claimed both documents as sources. A subsequent
per-document operation, acting on that now-incorrect claim, treated entities
reachable through the absorbed provenance as belonging to the document being
processed, and removed them.

The property merge did not delete anything by itself. It corrupted the
bookkeeping that a later, entirely correct routine depended on: which is why
the damage surfaced somewhere other than where it was caused, and why it was
found only by re-ingesting the affected document and noticing its entity count
had changed. We reconstructed this from the surviving graph state and the
provenance code path rather than from a trace of the original run, and flag it
as a reconstruction.

The precondition is still present by design (provenance accumulates, and
\S\ref{sec:identity} explains why it must) so the repair had to be
downstream, and it was. Per-document removal now deletes an item only when
withdrawing that document empties its provenance entirely: anything still backed
by another document simply drops the filename, and anything a human has edited
is retained with its provenance cleared instead. The repair narrows the blast radius rather than removing it, and it is worth
being exact about which. Items backed by another document are now safe: a
corrupted node's inflated provenance can no longer drag them out, which is the
class the eight belonged to only if they were also cited elsewhere, and we do
not know that they were. Items backed by the withdrawn document alone are still
removed with it (correctly, since nothing else vouches for them) so had the
same withdrawal run today, any of the eight that were single-sourced would still
have gone. What changed is that the decision now turns on the item's own
provenance rather than on a claim the merge fabricated.

\begin{figure}[H]
\centering
\fitwidth{%
\begin{tikzpicture}
 \node[ghdr] at (0,2.15) {record A: already in the graph};
 \node[grecok] (a) at (0,0.85) {\begin{tabular}{@{}l@{\hskip 5pt}l@{}}
  \texttt{name} & Sector Delta Power Grid\\
  \texttt{rdfType} & \texttt{PowerGridSector}\\
  \texttt{position} & its own\\
  \texttt{radius\_m} & n/a\\
  \texttt{is\_illustrative} & \texttt{false}\\
 \end{tabular}};

 \node[ghdr] at (0,-1.05) {record B: incoming, a different facility};
 \node[grec] (b) at (0,-2.35) {\begin{tabular}{@{}l@{\hskip 5pt}l@{}}
  \texttt{name} & Sector\_Delta\_Power\_Grid\\
  \texttt{rdfType} & \texttt{Location}\\
  \texttt{position} & another facility's\\
  \texttt{radius\_m} & \texttt{95000}\\
  \texttt{is\_illustrative} & \texttt{true}\\
 \end{tabular}};

 \node[gdiamond, text width=1.5cm] (m) at (5.4,-0.75) {string\\similarity};
 \node[gtiny, text width=2.2cm, align=center] at (4.9,-2.75)
    {identical modulo whitespace: the score was justified};

 \node[ghdr] at (10.2,1.25) {surviving node};
 \node[grecbad] (bad) at (10.2,-0.75) {\begin{tabular}{@{}l@{\hskip 5pt}l@{}}
  \texttt{name} & Sector Delta Power Grid\\
  \texttt{rdfType} & \textcolor{s5col}{\texttt{Location}}\\
  \texttt{position} & \textcolor{s5col}{another facility's}\\
  \texttt{radius\_m} & \textcolor{s5col}{\texttt{95000}}\\
  \texttt{is\_illustrative} & \textcolor{s5col}{\texttt{true}}\\
 \end{tabular}};

 \node[grecbad] (coll) at (10.2,-3.15) {\begin{tabular}{@{}l@{}}
  \textcolor{s5col}{8 entities of an unrelated document deleted}\\
 \end{tabular}};

 \draw[gflow] (a.east) -- (m);
 \draw[gflow] (b.east) -- (m);
 \draw[gflow] (m) -- (bad.west);
 \draw[gflow, dashed, draw=s5col!70] (bad.south) -- (coll.north);
\end{tikzpicture}}
\caption{The merge of \S\ref{sec:incident}. Both inputs were real records and
the similarity score was justified; the damage is entirely in the
property-combination step. No exception was raised at any point, and the
collateral deletion was found only on a later re-ingest.}
\label{fig:incident}
\end{figure}
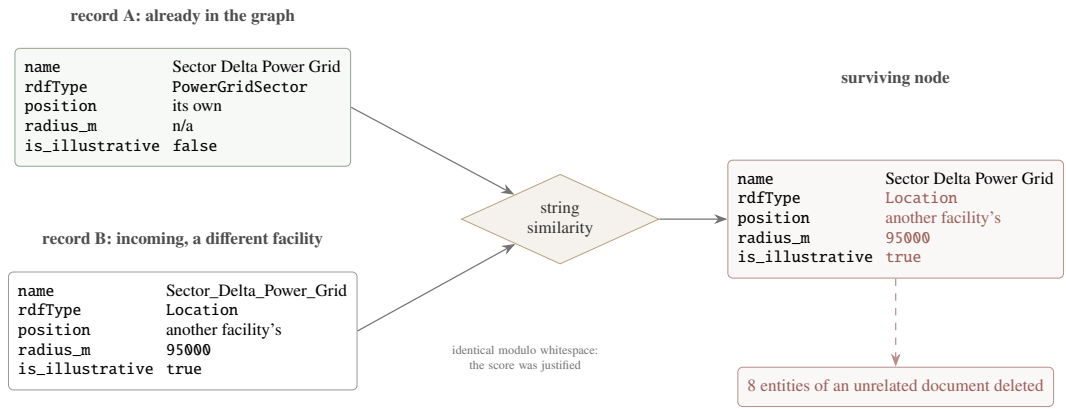

Three features of this failure shaped everything after it:

\begin{itemize}[leftmargin=*, itemsep=1pt, topsep=2pt]
 \item \textbf{No threshold would have prevented it.} The merge was
    high-confidence and the confidence was justified: the two names really
    were identical modulo whitespace. This is not a tuning failure.
 \item \textbf{It was silent.} No exception, no constraint violation, no
    warning. The corrupted node was structurally valid and looked plausible.
 \item \textbf{It was unrecoverable in place.} Once the property sets were
    combined, the information needed to split them no longer existed. The
    repair was manual and depended on the source documents.
\end{itemize}

The conclusion we draw is narrow and, we think, general: \emph{a destructive
operation whose errors are undetectable should not be automated on a similarity
signal, at any threshold}. Not "at a higher threshold": at any. The
similarity signal is measuring the wrong thing, and a better-calibrated version
of the wrong thing does not help. This applies to learned matchers as much as to
string metrics \cite{mudgal2018deep, li2020ditto}: they improve the matching
decision without changing what a wrong match costs, or whether anyone sees it.

\subsection{Embedding generation and backfill}
\label{sec:backfill}

Every node and edge carries an embedding \cite{reimers2019sbert} used by the
retrieval layer. When
embeddings must be generated for existing rows (after a model change, or for
rows written before the field existed) the work is a backfill over the whole
graph.

Our first backfill issued one asynchronous task per pending row: $1{,}427$
concurrent tasks against a driver with a bounded connection pool. The result was
a thundering herd in which tasks spent their time contending for connections
and retrying transactions rather than doing work. The run completed and reported
success. It had left approximately $7{,}000$ edges unembedded behind $1{,}790$
transaction retries; a second pass was required to clear them.

The defect worth naming is not the concurrency, which is easily bounded. It is
that \emph{the job reported a completion it had not achieved}. Each individual
task reported success or failure honestly, and the aggregate was computed from
the tasks rather than from the population. A backfill that never re-counts the
rows it was supposed to fill cannot know whether it filled them. We return to
this pattern in \S\ref{sec:pagination}, where it recurs in a different component
with the same shape.

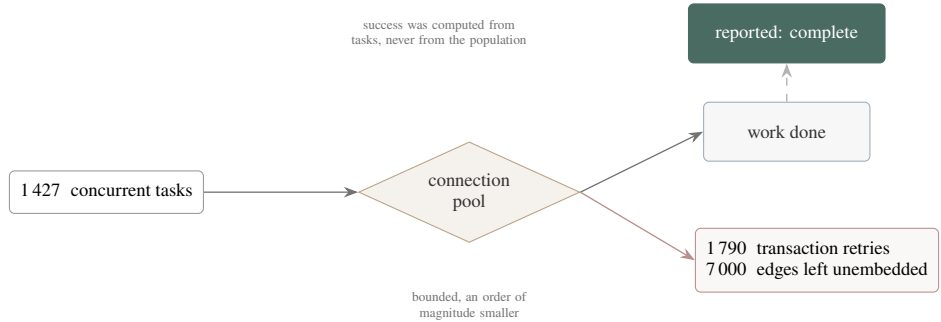
\begin{figure}[H]
\centering
\fitwidth{%
\begin{tikzpicture}
 \node[grec] (tasks) at (0,0) {\begin{tabular}{@{}r@{\hskip 4pt}l@{}}
  1\,427 & concurrent tasks\\
 \end{tabular}};
 \node[gdiamond, text width=1.6cm] (pool) at (4.8,0) {connection\\pool};
 \node[gtiny, text width=2.2cm, align=center] at (4.8,-1.5) {bounded, an order of\\magnitude smaller};

 \node[gproc, minimum width=2.2cm] (work) at (9.0,0.8) {work done};
 \node[grecbad] (retry) at (9.4,-0.9) {\begin{tabular}{@{}r@{\hskip 4pt}l@{}}
  1\,790 & transaction retries\\
  7\,000 & edges left unembedded\\
 \end{tabular}};

 \node[goutput, minimum width=2.6cm] (rep) at (9.0,2.1) {reported: complete};

 \draw[gflow] (tasks.east) -- (pool.west);
 \draw[gflow] (pool.east) -- (work.west);
 \draw[gflow, draw=s5col!70] (pool.east) -- (retry.west);
 \draw[gflow, dashed, draw=black!30] (work.north) -- (rep.south);
 \node[gtiny, text width=3.4cm, align=center] at (4.4,2.1)
    {success was computed from tasks, never from the population};
\end{tikzpicture}}
\caption{The backfill pathology. Each task reported its own outcome honestly;
the aggregate was derived from those reports rather than from a recount of
unembedded rows, so the run claimed a completion it had not reached.}
\label{fig:backfill}
\end{figure}

\subsection{The insertion path}
\label{sec:insertionpath}

Writing a document's extraction output into the graph was, in its first form,
slower than extracting it had been. The path from there to the present shape
went through four changes, each of which exposed the next bottleneck rather than
solving the problem outright.

\begin{table}[H]
\centering
\caption{How insertion cost fell. Most stages removed the dominant cost of the
previous one; the third did not, and is retained because a change that bought
nothing is what identified the real bound. Measurements come from the separate
evaluation corpus noted in \S\ref{sec:edgeid}.}
\label{tab:insertion}
\small
\begin{tabular}{@{}P{2.6cm}P{4.4cm}P{4.6cm}@{}}
\toprule
Stage & What was wrong & Effect \\
\midrule
Baseline &
 Extraction was re-run against content that had already been extracted &
 Over six hours for six documents \\
Direct insertion &
 The redundant second extraction removed entirely &
 A 45-page document fell from over an hour to 40--45 minutes \\
Concurrent embedding &
 Requests parallel but still one text each, so the call count was unchanged &
 Intermittent connection failures under load \\
Asynchronous persistence &
 Structure was waiting on embeddings that it did not need &
 Graph queryable one to two minutes after ingest \\
True batching &
 Twenty texts per call, five calls in flight, writes folded into \texttt{UNWIND} transactions &
 API calls and write transactions both fell from ${\sim}7{,}300$ to ${\sim}365$ \\
\bottomrule
\end{tabular}
\end{table}

The third row is the instructive one. Raising concurrency looked like the
obvious response to a slow embedding stage, and it did not help: the work was
bounded by the number of round trips, not by how many were in flight, so
twentyfold parallelism bought nothing and cost stability. Only packing several
texts into each request changed the quantity that actually governed the cost.
This is the same lesson as \S\ref{sec:backfill} from the opposite direction: there, concurrency was raised past what the pool could serve; here, it was raised
against a limit concurrency could not move.

% ══════════════════════════════════════════════════════════════
\section{Ontology tagging}
\label{sec:tagging}

\subsection{The ontology as a live database}
\label{sec:livedb}

Our model derives from BFO \cite{arp2015bfo} and CCO \cite{jensen2024cco},
extended with classes for Indian
governmental structures that the upper ontologies do not cover: an extension
under an upper-level commitment rather than a vocabulary of our own
\cite{smith2007obo}. It is not
parsed from Turtle files at start-up. It lives in a second Neo4j instance and
is read from there, which means a curator can add a class in the afternoon and
have the tagger use it that evening without a redeploy.

At the time of measurement that instance held $22{,}401$ triples: $5{,}173$
classes, $748$ named individuals, $7{,}600$ label entries once synonyms and
alternative labels are counted, and $849$ predicates carrying both a domain and
a range. These are overlapping counts over the same triples rather than a
partition of them: one class contributes several label entries, and a predicate
contributes both its domain and its range triple. The tagger loads all of it into memory once per run and builds a
normalised label index over it.

The arrangement has an obvious cost, which we mention because it caught us. The
tagger's picture of the ontology is exactly as current as its last load. A long
run tags its final entity against the vocabulary that existed when it started.
For runs measured in hours over half a million entities, that window is not
negligible, and it is the reason two tagging passes over the same graph can
disagree without either being wrong.

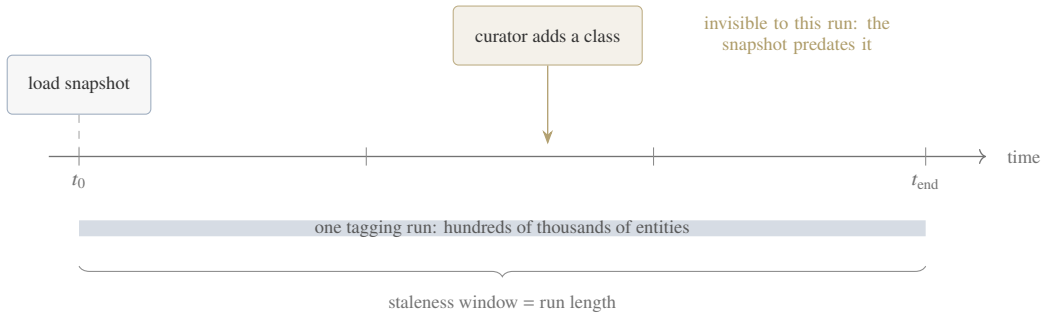
\begin{figure}[H]
\centering
\fitwidth{%
\begin{tikzpicture}[every node/.style={font=\scriptsize}]
 \draw[->, line width=0.5pt, draw=black!55] (0,0) -- (12.4,0);
 \node[text=black!55] at (12.9,0) {time};
 \foreach \x/\t in {0.4/{$t_0$}, 4.2/{}, 8.0/{}, 11.6/{$t_{\mathrm{end}}$}}
  \draw[line width=0.4pt, draw=black!45] (\x,0.12) -- (\x,-0.12) node[below, text=black!55] {\t};

 \node[gproc, minimum width=1.9cm] (load) at (0.4,0.95) {load snapshot};
 \draw[draw=black!35, line width=0.4pt, dashed] (0.4,0.55) -- (0.4,0.12);

 \draw[line width=6pt, draw=s1col!25] (0.4,-0.95) -- (11.6,-0.95);
 \node[text=black!65] at (6.0,-0.95) {one tagging run: hundreds of thousands of entities};

 \node[greview, minimum width=2.5cm] (edit) at (6.6,1.6) {curator adds a class};
 \draw[gflow, draw=bugcol!80] (edit.south) -- (6.6,0.16);
 \node[text=bugcol!85, align=center, text width=3.2cm] at (9.9,1.6)
    {invisible to this run: the snapshot predates it};

 \draw[decorate, decoration={brace, amplitude=4pt, mirror}, draw=black!45]
    (0.4,-1.45) -- (11.6,-1.45);
 \node[text=black!55] at (6.0,-1.95) {staleness window $=$ run length};
\end{tikzpicture}}
\caption{Why two passes over one graph can disagree without either being wrong.
The tagger reads the ontology once, at $t_0$; a class added during the run is
not seen by it. The window is the run length, which for a full pass over this
graph is measured in hours.}
\label{fig:staleness}
\end{figure}

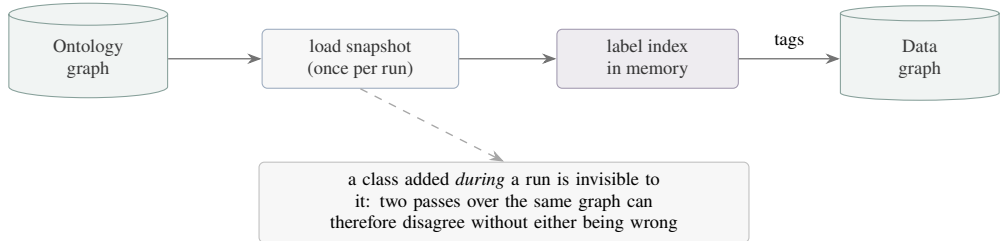
\begin{figure}[H]
\centering
\fitwidth{%
\begin{tikzpicture}
 \node[gdb] (odb) at (0,0) {Ontology\\graph};
 \node[gproc, minimum width=2.6cm] (load) at (3.6,0) {load snapshot\\(once per run)};
 \node[gonto, minimum width=2.4cm] (idx) at (7.4,0) {label index\\in memory};
 \node[gdb] (ddb) at (11.0,0) {Data\\graph};
 \node[gnote, text width=6.2cm] (note) at (5.5,-1.9)
    {\scriptsize a class added \emph{during} a run is invisible to it: two passes
     over the same graph can therefore disagree without either being wrong};
 \draw[gflow] (odb) -- (load);
 \draw[gflow] (load) -- (idx);
 \draw[gflow] (idx) -- node[above, font=\scriptsize] {tags} (ddb);
 \draw[gflow, dashed, draw=black!35] (load.south) -- (note.north);
\end{tikzpicture}}
\caption{The ontology is read from a second database rather than parsed from
files, so curation takes effect without a redeploy. The cost is a staleness
window equal to the run length.}
\label{fig:twodb}
\end{figure}

\subsection{Resolving a class for an entity}
\label{sec:resolve}

Given an entity with a name and an extractor-assigned type string, the tagger
looks for the most specific class that matches either
\cite{euzenat2013matching, shvaiko2013challenges}. Exact matching against
the normalised label index comes first \cite{mendes2011spotlight, shen2015entity}.
Where that fails, we generate variants (stripping parenthetical suffixes, splitting slash-separated compounds, and
attempting singular forms) and retry. Only if all of those miss do we fall
back to approximate string matching \cite{cheatham2013string} (a
longest-contiguous-matching-subsequence ratio, thresholded at $0.9$) deliberately high, because the failure mode of a loose fuzzy match is a
confident wrong answer rather than an obvious one. That cutoff, and the $0.90$
and $0.95$ cosine thresholds of \S\ref{sec:agent}, were chosen from the
behaviour of the existing canonicalisation code rather than tuned against a
labelled set; we have no evidence that they are optimal.

When several classes match, we prefer the more specific. Specificity is taken
from depth in the subsumption hierarchy, with a preference for classes carrying
a segment tag matching the document's own segment, so that an ambiguous label
resolves differently in a naval document than in a municipal one.

\begin{figure}[H]
\centering
\fitwidth{%
\begin{tikzpicture}
 \node[ghdr] at (0,1.5) {input};
 \node[grec] (in) at (0,0.6) {\begin{tabular}{@{}l@{\hskip 5pt}l@{}}
  \texttt{name} & Minehunter (MHC)\\
  \texttt{type} & vessels\\
 \end{tabular}};

 \node[gproc, minimum width=3.0cm] (s1) at (4.9,1.65) {exact index lookup};
 \node[gtiny, text width=3.6cm, align=left] at (9.4,1.65)
    {\texttt{"minehunter (mhc)"} $\to$ miss\\\texttt{"vessels"} $\to$ miss};

 \node[gproc, minimum width=3.0cm] (s2) at (4.9,0.15) {variant generation};
 \node[gtiny, text width=3.6cm, align=left] at (9.4,0.15)
    {parenthetical $\to$ \texttt{"mhc"}\\stripped $\to$ \texttt{"minehunter"}\\singular $\to$ \texttt{"vessel"}};

 \node[gproc, minimum width=3.0cm] (s3) at (4.9,-1.45) {fuzzy, cutoff $0.9$};
 \node[gtiny, text width=3.6cm, align=left] at (9.4,-1.45)
    {only if every variant misses;\\a loose match fails confidently};

 \node[gproc, minimum width=3.0cm] (s4) at (4.9,-2.95) {rank the survivors};
 \node[gtiny, text width=3.6cm, align=left] at (9.4,-2.95)
    {deeper class wins; segment tag\\breaks a remaining tie};

 \node[goutput, minimum width=3.0cm] (out) at (0,-2.95) {\texttt{MineHunter}};
 \node[gtiny, text width=2.8cm, align=center] at (0,-4.05)
    {type string also yields \texttt{Vessel}, kept as a role by Eq.~\ref{eq:roles}};

 \draw[gflow] (in.east) -- (s1.west);
 \draw[gflow] (s1) -- (s2);
 \draw[gflow] (s2) -- (s3);
 \draw[gflow] (s3) -- (s4);
 \draw[gflow] (s4.west) -- (out.east);
 \draw[gflow, dashed, draw=black!30] (s2.east) -- (7.6,0.15);
\end{tikzpicture}}
\caption{Class resolution for one entity. Each stage is attempted only when the
previous one misses, so the expensive and least reliable step (approximate
matching) runs last and against a high cutoff. The parenthetical rule that
recovers \texttt{MHC} here is the same rule that misfires on qualifiers
(\S\ref{sec:asymmetry}).}
\label{fig:resolve}
\end{figure}
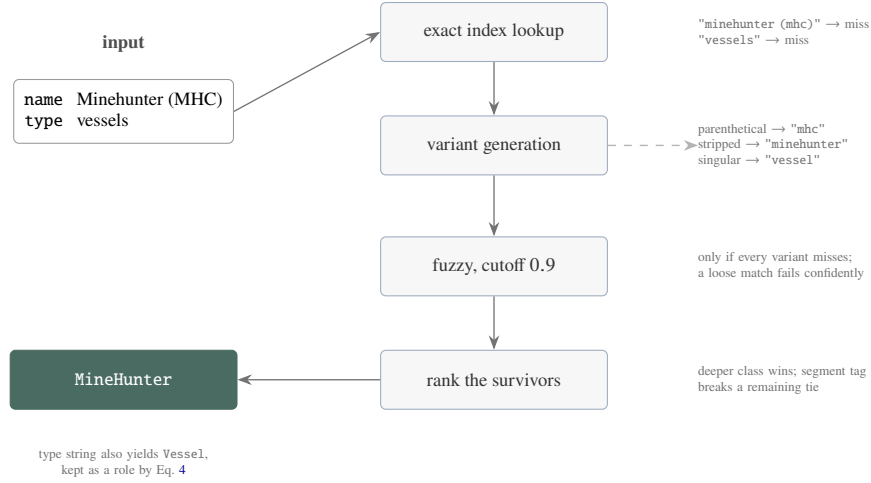

\subsection{Multi-class tagging}
\label{sec:multiclass}

An entity often belongs to more than one class at once, and those classes are
frequently unrelated by subsumption \cite{tsoumakas2007multilabel}. A named
official is a \emph{Person} and
also holds an office: a role rather than a kind
\cite{steimann2000roles, loebe2007roles}. A depot is a \emph{Depot} and, in an
operational context,
a vulnerable point. Recording only the winner of that contest throws away
whichever fact the next query happens to need.

So the tagger keeps them all. The best match is stored as the primary type, for
compatibility with everything downstream that assumes one; the remainder are
stored in a field named \texttt{roles}. That name is ours and predates our
reading of the roles literature; it denotes any additional class an entity
carries, which in the sense of Steimann and Loebe is usually a \emph{kind}
rather than a role: \texttt{GovernmentOrganization} is something an entity
is, not a part it plays. We keep the field name for continuity with the
implementation and disclaim the theoretical reading. A candidate survives only if it is genuinely orthogonal to the
primary:

\begin{equation}
\mathcal{R}(e) = \bigl\{\, c \in \mathcal{C}(e) \;:\; c \neq p,\;
 \neg\,(c \sqsubseteq p) \;\wedge\; \neg\,(p \sqsubseteq c) \,\bigr\},
\qquad p = \arg\max_{c \in \mathcal{C}(e)} \mathrm{spec}(c, \sigma)
\label{eq:roles}
\end{equation}

Specificity is written $\mathrm{spec}(c,\sigma)$ because it is relative to the
document's segment $\sigma$: depth in the hierarchy settles most cases and a
class carrying the document's own segment tag settles the rest.

The subsumption test in Equation~\ref{eq:roles} matters more than it looks. An
ancestor of the primary adds nothing, because the hierarchy above the primary is
already reachable from it and is rendered separately as the is-a chain
\cite{brachman1983isa}; a
descendant adds nothing for the same reason in the other direction. Without
this filter the role list fills with restatements of the primary and the genuine
cross-cutting classifications are lost in them.

The filter is only as good as the axioms it consults, and this cuts both ways.
An entity typed \texttt{MinistryOfFinance} also carries
\texttt{GovernmentOrganization} as a role, which Equation~\ref{eq:roles} admits
because the ontology's chain for the former runs through
\texttt{Organization} and \texttt{GroupOfAgents} without passing through the
latter. Read as orthogonality that is correct; read as modelling it is at least
arguable, since a ministry of finance is plainly a government organisation. Where
a subsumption axiom that ought to exist is missing, a pair that should have been
collapsed is instead recorded as two independent classifications. Some unknown
fraction of the multi-class population is therefore measuring gaps in the
ontology rather than genuine cross-cutting structure, and we have no way to
separate the two without an axiom-completeness audit we have not performed.

\begin{figure}[H]
\centering
\fitwidth{%
\begin{tikzpicture}
 \node[ghdr] at (0,1.45) {entity node as stored};
 \node[grec] (n) at (0,0) {\begin{tabular}{@{}l@{\hskip 5pt}l@{}}
  \texttt{name} & Ministry of Finance\\
  \texttt{rdfType} & \texttt{oas:MinistryOfFinance}\\
  \texttt{rdfTypeLabel} & MinistryOfFinance\\
  \texttt{roles} & \texttt{[oas:GovernmentOrganization]}\\
  \texttt{roleLabels} & \texttt{["Government Organization"]}\\
 \end{tabular}};

 \node[ghdr] at (9.2,1.45) {reachable by traversal, so not stored};
 \node[gdistinct, minimum width=4.6cm] (c1) at (9.2,0.55) {\texttt{oas:Organization}};
 \node[gdistinct, minimum width=4.6cm] (c2) at (9.2,-0.25) {\texttt{cco:GroupOfAgents}};
 \node[gdistinct, minimum width=4.6cm] (c3) at (9.2,-1.05) {\texttt{bfo:independent continuant}};
 \draw[gflow, draw=black!30] (c1) -- (c2);
 \draw[gflow, draw=black!30] (c2) -- (c3);

 \draw[gflow, dashed] (n.east) -- node[above, font=\scriptsize, text=black!55] {\texttt{subClassOf}$^{*}$} (c1.west);
 \node[gtiny, text width=5.0cm, align=center] at (4.6,-2.2)
    {Equation~\ref{eq:roles} drops any candidate related to the primary by
    subsumption in either direction; only orthogonal classes reach
    \texttt{roles}};
\end{tikzpicture}}
\caption{An entity carrying two orthogonal classes. The is-a chain above the
primary is reachable by traversal and is therefore excluded from the role set
by Equation~\ref{eq:roles}; only classifications that the hierarchy does not
already imply are stored. The pairing shown is admissible under
Equation~\ref{eq:roles} because the ontology places \texttt{MinistryOfFinance}
beneath \texttt{Organization} and \texttt{GroupOfAgents}, not beneath
\texttt{GovernmentOrganization}: the two are siblings rather than related by
subsumption, so the filter correctly retains the second. Whether they
\emph{should} be siblings is the separate question raised in
\S\ref{sec:multiclass}.}
\label{fig:multiclass}
\end{figure}
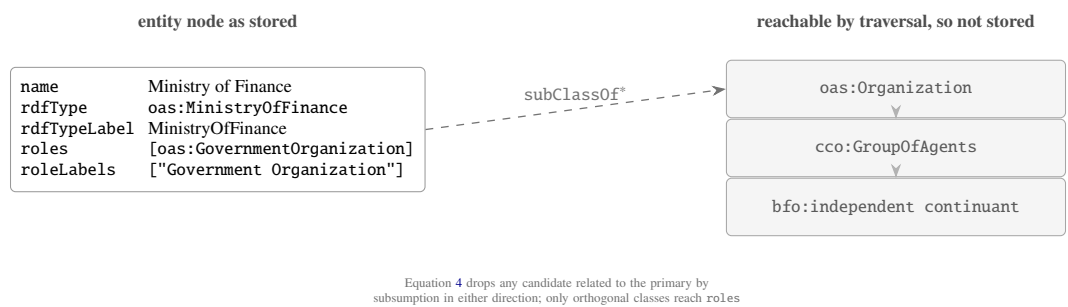

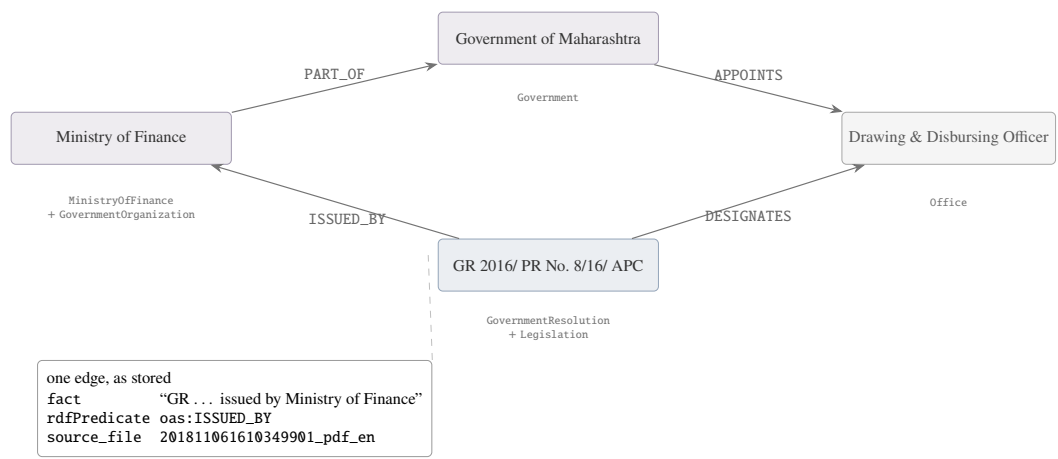
\begin{figure}[H]
\centering
\fitwidth{%
\begin{tikzpicture}
 \node[gonto, minimum width=3.3cm] (mof) at (0,0) {Ministry of Finance};
 \node[gtiny, text width=3.4cm, align=center] at (0,-1.05)
    {\texttt{MinistryOfFinance}\\$+$ \texttt{GovernmentOrganization}};

 \node[gonto, minimum width=3.3cm] (gom) at (6.4,1.5) {Government of Maharashtra};
 \node[gtiny, text width=2.8cm, align=center] at (6.4,0.62) {\texttt{Government}};

 \node[groute, minimum width=3.3cm] (gr) at (6.4,-1.9) {GR 2016/ PR No.\ 8/16/ APC};
 \node[gtiny, text width=3.2cm, align=center] at (6.4,-2.85)
    {\texttt{GovernmentResolution}\\$+$ \texttt{Legislation}};

 \node[gdistinct, minimum width=3.0cm] (off) at (12.4,0) {Drawing \& Disbursing Officer};
 \node[gtiny, text width=2.8cm, align=center] at (12.4,-0.95) {\texttt{Office}};

 \draw[gflow] (mof) -- node[above, font=\scriptsize, text=black!60] {\texttt{PART\_OF}} (gom);
 \draw[gflow] (gr) -- node[below, font=\scriptsize, text=black!60, pos=0.45] {\texttt{ISSUED\_BY}} (mof);
 \draw[gflow] (gom) -- node[above, font=\scriptsize, text=black!60] {\texttt{APPOINTS}} (off);
 \draw[gflow] (gr) -- node[below, font=\scriptsize, text=black!60] {\texttt{DESIGNATES}} (off);

 \node[grec] (prov) at (1.7,-4.05) {\begin{tabular}{@{}l@{\hskip 4pt}l@{}}
  \multicolumn{2}{@{}l@{}}{one edge, as stored}\\[1pt]
  \texttt{fact} & ``GR \dots\ issued by Ministry of Finance''\\
  \texttt{rdfPredicate} & \texttt{oas:ISSUED\_BY}\\
  \texttt{source\_file} & \texttt{201811061610349901\_pdf\_en}\\
 \end{tabular}};
 \draw[draw=black!25, line width=0.4pt, dashed] (prov.north east) -- (4.6,-1.75);
\end{tikzpicture}}
\caption{An excerpt of the graph as built. Classes appear beneath each entity;
two of the four carry a second, non-subsuming class. Every edge is typed with an
ontology predicate and retains every document that asserted it (the singular
\texttt{source\_file} holds the last writer, while \texttt{source\_files}
accumulates all of them) which is what allows an answer to be traced back to
its pages, including the corroborating ones.}
\label{fig:excerpt}
\end{figure}

\subsection{The name-versus-type evidence asymmetry}
\label{sec:asymmetry}

The tagger has two sources of evidence, and we initially treated them as
interchangeable. They are not, and the difference is the most useful thing we
learned building this layer.

A type string is a type assertion. Something (a model, a schema, a human) has claimed that this entity is of that kind. An entity's name is not an
assertion of anything. It is a label attached to an individual, and the fact
that a type word happens to appear inside it carries no claim at all. The
distinction is the class/instance one, and conflating the two is a known way to
build a taxonomy that does not mean what it says
\cite{brachman1983isa, guarino2002ontoclean}.

The distinction is invisible until you look at what it produces. "Government
Resolution No.~30/2013" announces its own type; taking \emph{Government
Resolution} from it is right, in the way a lexico-syntactic pattern is right
\cite{hearst1992hyponyms}. "Under Secretary (Regulations)" announces an
office, and the parenthetical qualifies which office. Extracting
\emph{Regulations} from it and matching that against the class index yields
\emph{Statutory Rule} and \emph{Legislation}, so a person ends up classified as
legislation. Our variant generator was doing exactly this, and doing it
correctly by its own specification: the parenthetical rule exists to turn
"Minehunter (MHC)" into the abbreviation MHC, which is a real and useful case.
It simply has no way to tell an abbreviation from a qualifier.

We now admit a name-derived candidate only when the class label is anchored in
the name rather than merely present in it, which makes name evidence a weaker
signal than an asserted type rather than an equal one
\cite{paulheim2013type}. Writing $\mathrm{tok}(\cdot)$ for the
normalised token sequence, a candidate class $c$ is admissible for name $n$ when

\begin{equation}
\mathrm{tok}(\ell_c) \sqsubseteq_{\mathrm{pre}} \mathrm{tok}(n)
\;\;\vee\;\;
\bigl(|\mathrm{tok}(\ell_c)| \ge 2 \;\wedge\;
   \mathrm{tok}(\ell_c) \sqsubseteq_{\mathrm{sub}} \mathrm{tok}(n)\bigr)
\label{eq:evidence}
\end{equation}

where $\sqsubseteq_{\mathrm{pre}}$ is the prefix relation and
$\sqsubseteq_{\mathrm{sub}}$ contiguous subsequence. In words: the class label
either opens the name, or appears inside it as a phrase of at least two tokens.
A single token is admitted only in the first case: a one-word label that
\emph{begins} the name still qualifies, so ``Ministry of Finance'' does admit
\emph{Ministry}. The two-token floor constrains interior matches only, which is
where the parenthetical failures of this section arise; we state this because
the rule is weaker than a reader might infer, and every removal we observed came
from an interior match. Type-derived candidates bypass
Equation~\ref{eq:evidence} entirely, since they are assertions and not
inferences.

The two-token floor is doing most of the work. Reviewing the assignments the
rule removed, all $32$ were single generic words (Information, Programme,
Ministry, Channel, Project) picked out of parentheticals; the four assignments
it kept were the four that manual review had confirmed correct. The ones
it kept were multi-word phrases that genuinely occurred in the name, such as
\emph{Gazetted Officer} inside "(Group-A) Gazetted Officer". An earlier and
stricter version of the rule required the label to \emph{open} the name, and it
discarded that example along with the junk; the anchoring disjunction in
Equation~\ref{eq:evidence} exists because of it.

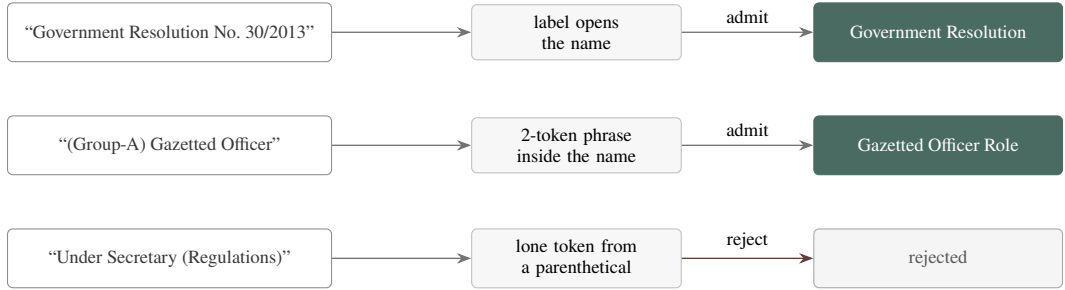
\begin{figure}[H]
\centering
\fitwidth{%
\begin{tikzpicture}
 \node[gdata, minimum width=4.3cm] (n1) at (0,0)   {\scriptsize ``Government Resolution No.~30/2013''};
 \node[gdata, minimum width=4.3cm] (n2) at (0,-1.5) {\scriptsize ``(Group-A) Gazetted Officer''};
 \node[gdata, minimum width=4.3cm] (n3) at (0,-3.0) {\scriptsize ``Under Secretary (Regulations)''};

 \node[gnote, text width=2.5cm] (t1) at (5.4,0)  {\scriptsize label opens\\the name};
 \node[gnote, text width=2.5cm] (t2) at (5.4,-1.5) {\scriptsize 2-token phrase\\inside the name};
 \node[gnote, text width=2.5cm] (t3) at (5.4,-3.0) {\scriptsize lone token from\\a parenthetical};

 \node[goutput,  minimum width=3.3cm] (o1) at (10.2,0)  {\scriptsize Government Resolution};
 \node[goutput,  minimum width=3.3cm] (o2) at (10.2,-1.5) {\scriptsize Gazetted Officer Role};
 \node[gdistinct, minimum width=3.3cm] (o3) at (10.2,-3.0) {\scriptsize rejected};

 \foreach \a/\b in {n1/t1, n2/t2, n3/t3} \draw[gflow] (\a) -- (\b);
 \draw[gflow] (t1) -- node[above, font=\scriptsize] {admit} (o1);
 \draw[gflow] (t2) -- node[above, font=\scriptsize] {admit} (o2);
 \draw[gflow, draw=s5col!65!black] (t3) -- node[above, font=\scriptsize] {reject} (o3);
\end{tikzpicture}}
\caption{Equation~\ref{eq:evidence} on three real names. The third would have
classified a person as \emph{Statutory Rule} and \emph{Legislation}; the second
is why the rule permits an interior phrase rather than requiring a prefix.}
\label{fig:evidence-cases}
\end{figure}

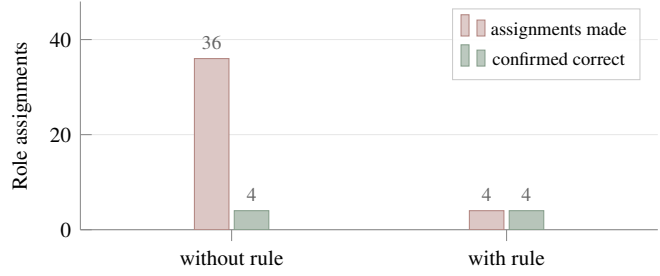
\begin{figure}[H]
\centering
\fitwidth{%
\begin{tikzpicture}
\begin{axis}[
 width=0.66\textwidth, height=4.6cm,
 ybar, bar width=13pt,
 ymin=0, ymax=48,
 ylabel={Role assignments},
 legend style={at={(0.97,0.97)}, anchor=north east, font=\scriptsize,
        draw=black!20, fill=white, fill opacity=0.92, text opacity=1, line width=0.4pt},
 ylabel style={font=\footnotesize},
 symbolic x coords={without rule, with rule},
 xtick=data,
 xticklabel style={font=\footnotesize},
 yticklabel style={font=\footnotesize},
 nodes near coords, nodes near coords style={font=\scriptsize, text=black!60},
 enlarge x limits=0.55,
 axis lines*=left,
 axis line style={draw=black!45, line width=0.4pt},
 tick style={draw=black!45, line width=0.4pt},
 ymajorgrids, grid style={black!10, line width=0.3pt},
]
\addplot+[draw=s5col!75, fill=s5col!35] coordinates {(without rule,36) (with rule,4)};
\addplot+[draw=s4col!75, fill=s4col!45] coordinates {(without rule,4) (with rule,4)};
\legend{\scriptsize assignments made, \scriptsize confirmed correct}
\end{axis}
\end{tikzpicture}}
\caption{Effect of Equation~\ref{eq:evidence} on 2{,}500 sampled entities. Every
assignment surviving the rule was manually checked and found correct; the 32
removed were single generic tokens taken from parentheticals. The sample was
drawn by taking the first 4{,}000 entities the store returned and selecting
2{,}500 of them at random, which is not a uniform draw over the graph: it carries $1.44$ role
assignments per hundred entities pre-rule against roughly $0.72$ per hundred
graph-wide (the graph-wide figure converted using the bounded ratio of
\S\ref{sec:asymmetry}, so it is a lower bound), so it over-represents the phenomenon about twofold. Post-rule the two
agree ($0.16$ against $0.18$ per hundred). Both figures are assignments per
entity rather than entity counts, which are not interchangeable. The precision result should be read as measured on
a sample enriched for the phenomenon, which is the right sample for judging
precision and the wrong one for estimating prevalence.}
\label{fig:evidence}
\end{figure}

Applied across the whole graph, the rule cut entities carrying two or more
classes from $3{,}807$ to $989$. We first read that as over-correction, then as
a unit mismatch. Neither survives the arithmetic, and the honest position is that
part of the gap is still unexplained.

\begin{figure}[H]
\centering
\fitwidth{%
\begin{tikzpicture}
\begin{axis}[
 width=0.66\textwidth, height=4.4cm,
 ybar, bar width=14pt,
 ymin=1, ymax=5000000,
 ymode=log, log origin=infty,
 ylabel={Entities (log scale)},
 scaled y ticks=false,
 /pgf/number format/1000 sep={,},
 ylabel style={font=\footnotesize},
 xlabel={Ontology classes carried}, xlabel style={font=\footnotesize},
 symbolic x coords={1, 2, 3},
 xtick=data,
 xticklabel style={font=\footnotesize}, yticklabel style={font=\footnotesize},
 point meta=explicit symbolic,
 nodes near coords, nodes near coords style={font=\scriptsize, text=black!60},
 enlarge x limits=0.4,
 axis lines*=left,
 axis line style={draw=black!45, line width=0.4pt},
 tick style={draw=black!45, line width=0.4pt},
 ymajorgrids, grid style={black!10, line width=0.3pt},
]
\addplot+[draw=ontocol!75, fill=ontocol!18] coordinates
 {(1,503769) [503{,}769] (2,2826) [2{,}826] (3,36) [36]};
\end{axis}
\end{tikzpicture}}
\caption{Distribution of class count per entity, with raw counts labelled on a
logarithmic axis. \textbf{Captured mid-pass}, during the re-tagging run of
\S\ref{sec:asymmetry}: the $2{,}862$ entities shown here with two or more
classes sit between the $3{,}807$ that held them before the evidence rule was
applied and the $989$ of Table~\ref{tab:coverage} once the pass completed. The
shape is the point (multi-class assignment is rare but the tail is real, and
nothing in the representation caps the count) no entity carried four classes
at this snapshot, though one has since been given four by hand.}
\label{fig:classdist}
\end{figure}
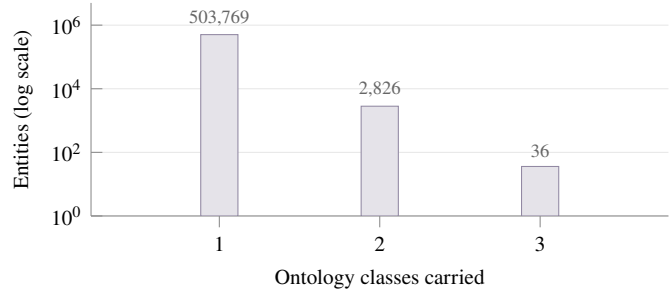

The two rates are in different units (assignments in the sample, entities
holding two or more classes graph-wide) and converting between them needs a
roles-per-entity ratio we can only estimate. Figure~\ref{fig:classdist} offers
one: of the $2{,}862$ multi-class entities in that snapshot, $2{,}826$ hold one
role and $36$ hold two, giving $1.013$. At that ratio $3{,}807$ entities
correspond to about $3{,}855$ assignments and $989$ to about $1{,}001$, an
assignment-level fall of $74.0\%$: indistinguishable from the entity-level
fall.

That estimate is weaker than it looks, and in a direction that matters. The
snapshot was taken mid-pass, so it blends entities the run had already reached
with entities it had not. Since the rule removes roles it also flattens the
distribution, which makes $1.013$ an upper bound on the post-rule ratio and a
\emph{lower} bound on the pre-rule one. Using the low figure for the pre-rule
population understates pre-rule assignments, and so understates the
assignment-level fall: precisely the direction that would shrink the gap we
are trying to explain. A pre-rule ratio of $1.3$, for instance, would lift the
fall to about $80\%$ and halve the residual; closing it entirely would need
roughly $2.4$ roles per multi-class entity, which we consider implausible.

The honest statement is therefore a bound rather than a value: the sample's
$89\%$ exceeds the graph's fall by \emph{at most} about fifteen points, and by
an unknown amount less. Recovering the exact figure needs the class-count
distributions from the passes that produced $3{,}807$ and $989$ separately,
which we did not retain. What the sign does establish is that the graph removed
proportionally fewer assignments than the sample, not more: so there is no
evidence here of over-correction in either reading.

The likeliest explanation is in our own sampling. As
Figure~\ref{fig:evidence}'s caption records, the $2{,}500$ entities were not
drawn uniformly and over-represent multi-class entities roughly twofold; a
sample enriched for the phenomenon is plausibly also enriched for the junk
parenthetical roles the rule is built to remove, which would raise its removal
rate above the graph's. We offer that as the provisional account rather than a
demonstrated one: confirming it needs a uniform draw, which we did not take.

We had also suspected the implementation of filtering type-derived candidates,
which Equation~\ref{eq:evidence} exempts. Reading the code settles it: the
admissibility test is applied only to candidates that originate from the name,
and type-derived candidates bypass it entirely, as specified. What remains open
is the recall question in its own terms (whether any \emph{correct}
name-derived classifications were lost) and that needs a graph-wide count of
assignments before and after, not the entity counts we have.

\subsection{Coverage}

\begin{table}[H]
\centering
\caption{Tagging coverage at the point of measurement. The relationship pass was
interrupted by an infrastructure outage and is not complete; the entity figures
are.}
\label{tab:coverage}
\small
\begin{tabular}{lrrr}
\toprule
Population & Tagged & Total & Share \\
\midrule
Entities with a primary class      & 506{,}631   & 537{,}157   & 94.3\% \\
Entities with $\ge 2$ classes      & 989      & 537{,}157   & 0.2\% \\
Relationships with a predicate      & 1{,}141{,}443 & 2{,}198{,}567 & 51.9\% \\
\bottomrule
\end{tabular}
\end{table}

The $30{,}526$ entities without any class are not a tagger failure so much as an
ontology gap: they are names for which no class exists at any level of
specificity. They are the raw material for the proposal queue of
\S\ref{sec:queue}.

\subsection{A silent-correctness failure in the tagger}
\label{sec:pagination}

The tagger walks the untagged population in batches, and the obvious way to
write that walk is with an offset that advances by the batch size. That is
wrong here, and it was wrong in our code for some time.

The selector is \emph{untagged entities}. Tagging a row removes it from the
selector. So a batch that successfully tags $m$ of its $b$ rows shrinks the
population by $m$, and an offset advanced by the full $b$ skips $m$ rows that
were never examined. Over a long run this compounds: roughly half the population
went untouched while the job logged steady progress and exited reporting
success.

The fix is to advance only past the rows that stayed in the selector, since
those are the ones the next query will see again:

\begin{equation}
\mathrm{offset}_{k+1} =
\begin{cases}
\mathrm{offset}_k + b, & \text{when re-tagging everything (selector is stable)}\\[2pt]
\mathrm{offset}_k + \bigl(b - m_k\bigr), & \text{when selecting only untagged rows}
\end{cases}
\end{equation}

This is the same defect as the backfill of \S\ref{sec:backfill} wearing
different clothes. In both cases every individual operation succeeded, every
log line was accurate, and the aggregate claim was false, because completion was
inferred from work attempted rather than measured against the population. We
have started treating "did the count of remaining items actually reach zero" as
the only acceptable termination condition for any batch job over this graph.

\begin{figure}[H]
\centering
\fitwidth{%
\begin{tikzpicture}[every node/.style={font=\scriptsize}]
 \node[anchor=east] at (-0.55,0) {pass 1};
 \node[draw=black!35, fill=s4col!30, minimum width=0.9cm, minimum height=0.55cm] at (0.00,0) {1};
 \node[draw=black!35, fill=cardbg,  minimum width=0.9cm, minimum height=0.55cm] at (1.05,0) {2};
 \node[draw=black!35, fill=s4col!30, minimum width=0.9cm, minimum height=0.55cm] at (2.10,0) {3};
 \node[draw=black!35, fill=cardbg,  minimum width=0.9cm, minimum height=0.55cm] at (3.15,0) {4};
 \node[draw=black!35, fill=cardbg,  minimum width=0.9cm, minimum height=0.55cm] at (4.20,0) {5};
 \node[draw=black!35, fill=cardbg,  minimum width=0.9cm, minimum height=0.55cm] at (5.25,0) {6};
 \node[draw=black!35, fill=cardbg,  minimum width=0.9cm, minimum height=0.55cm] at (6.30,0) {7};
 \node[draw=black!35, fill=cardbg,  minimum width=0.9cm, minimum height=0.55cm] at (7.35,0) {8};
 \draw[draw=s1col!70, line width=0.5pt, rounded corners=2pt] (-0.52,-0.42) rectangle (3.67,0.42);
 \node[text=s1col!75] at (1.55,0.78) {\textsc{skip} 0, \textsc{limit} 4: rows 1 and 3 tagged};

 \node[anchor=east] at (-0.55,-1.9) {pass 2};
 \node[draw=black!35, fill=cardbg, minimum width=0.9cm, minimum height=0.55cm] at (0.00,-1.9) {2};
 \node[draw=black!35, fill=cardbg, minimum width=0.9cm, minimum height=0.55cm] at (1.05,-1.9) {4};
 \node[draw=black!35, fill=cardbg, minimum width=0.9cm, minimum height=0.55cm] at (2.10,-1.9) {5};
 \node[draw=black!35, fill=cardbg, minimum width=0.9cm, minimum height=0.55cm] at (3.15,-1.9) {6};
 \node[draw=black!35, fill=cardbg, minimum width=0.9cm, minimum height=0.55cm] at (4.20,-1.9) {7};
 \node[draw=black!35, fill=cardbg, minimum width=0.9cm, minimum height=0.55cm] at (5.25,-1.9) {8};
 \draw[draw=s5col!70, line width=0.5pt, rounded corners=2pt] (3.68,-2.32) rectangle (5.77,-1.48);
 \node[text=s5col!75!black] at (4.72,-2.68) {\textsc{skip} 4};
 \draw[line width=0.5pt, draw=s5col!70] (1.62,-2.32) -- (3.62,-2.32);
 \node[text=s5col!75] at (2.62,-2.68) {5 and 6 never examined};
\end{tikzpicture}}
\caption{Why a fixed offset loses rows. Tagging removes rows from the selector,
so the population shifts left between passes while the offset advances by the
full batch size. Rows 5 and 6 are skipped, and the job still reports success.}
\label{fig:skip}
\end{figure}
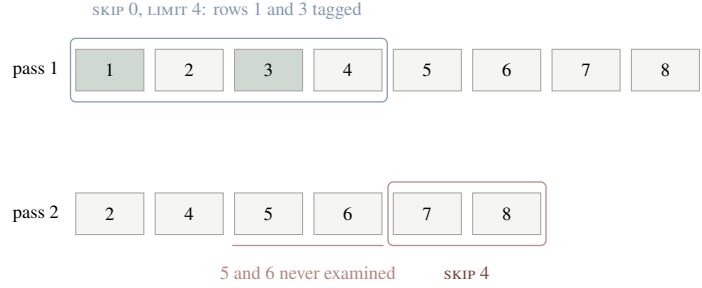

% ══════════════════════════════════════════════════════════════
\section{Conformance debt}
\label{sec:conformance}

Tagging assigns classes and predicates; it does not verify that the resulting
triples respect the ontology's own constraints: which is what makes an
ontology a specification and not merely a vocabulary
\cite{gruber1993translation}, and the check that constraint languages exist to
mechanise \cite{knublauch2017shacl}. Running that check over the
graph gave a lopsided picture. Entity typing was clean: no entity carried a
class absent from the ontology. The relationship layer was not. The
validator distinguishes three faults and reports them separately, because they
fail for different reasons.

\begin{table}[H]
\centering
\caption{Conformance faults on the production graph. Captured in a single
validator run and pending re-measurement (\S\ref{sec:limits}).}
\label{tab:debt-counts}
\small
\begin{tabular}{@{}lrP{5.8cm}@{}}
\toprule
Fault & Count & Meaning \\
\midrule
\textsc{unknown\_type} & 0 & An entity carries a class the ontology does not define \\
\textsc{unknown\_predicate} & 11{,}022 & A relationship uses a predicate the ontology does not define \\
\textsc{domain\_violation} & 10{,}025 & The source entity's type is not permitted for that predicate \\
\textsc{range\_violation} & 12{,}407 & The target entity's type is not permitted \\
\bottomrule
\end{tabular}
\end{table}

A domain violation usually means the subject was typed wrongly; a range
violation more often means the predicate itself was specified too loosely. Kept
apart, the two point at different repairs.

We give the counts in a table for legibility, but they should be read as one
qualitative finding (relation phrasing sprawls where entity typing does not) rather than as three quantitative results. Three qualifications, all of
which cut against reading them as settled measurements. First, they are \emph{floors}. Only $51.9\%$ of relationships
carried a predicate when the validator ran (Table~\ref{tab:coverage}), and an
untyped edge can register none of the three faults (not a domain or range
violation, and not an undefined predicate either) so all three counts are floors. If the untyped
remainder behaves like the typed portion, completing the pass should roughly
double each of the three. Second, the counts are
occurrences rather than distinct predicates; for canonicalisation (the repair
these numbers argue for) the actionable quantity is how many distinct
undefined predicates there are, which we did not record separately and should
have. Third, \textsc{unknown\_type} $=0$ is closer to a construction guarantee
than a quality result: the tagger can only assign classes it read out of the
ontology graph, so the count is zero by the way tagging works. It would become
non-zero if a curator deleted a class after a tagging run, which the staleness
window of \S\ref{sec:livedb} makes possible. Debt is the right word for the result: none of it
blocks a query today, and all of it raises the cost of every later change
\cite{sculley2015debt, zaveri2016quality, paulheim2017refinement}.

The asymmetry has a simple cause. Entity types are drawn from a curated,
enumerable vocabulary, and when extraction proposes something outside it the
mismatch is obvious. Relationship phrasing is open-ended
\cite{banko2007openie, fader2011reverb}: a model asked to name
the relation between an authority and a beneficiary will produce
\textsc{sanctions\_loan\_to}, \textsc{loan\_sanctioned\_by} and
\textsc{grants\_loan\_to} across three documents, all reasonable, none
identical, and each a new predicate as far as the graph is concerned.

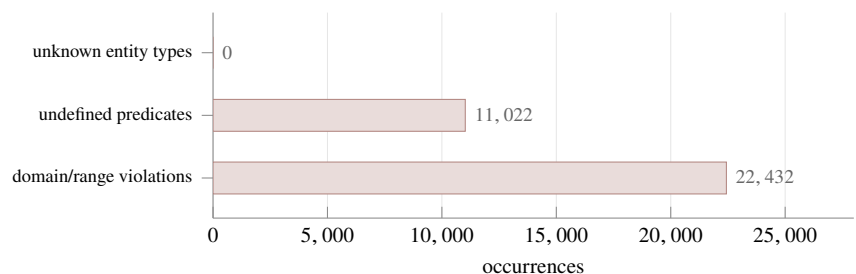
\begin{figure}[H]
\centering
\begin{tikzpicture}
\begin{axis}[
 width=0.72\textwidth, height=4.3cm,
 xbar, bar width=12pt,
 xmin=0, xmax=28000,
 xlabel={occurrences}, xlabel style={font=\footnotesize},
 symbolic y coords={domain/range violations, undefined predicates, unknown entity types},
 ytick=data,
 yticklabel style={font=\scriptsize}, xticklabel style={font=\footnotesize},
 nodes near coords, nodes near coords style={font=\scriptsize, text=black!60},
 scaled x ticks=false, /pgf/number format/1000 sep={,},
 enlarge y limits=0.32,
 axis lines*=left,
 axis line style={draw=black!45, line width=0.4pt},
 tick style={draw=black!45, line width=0.4pt},
 xmajorgrids, grid style={black!10, line width=0.3pt},
]
\addplot+[draw=s5col!75, fill=s5col!22] coordinates
 {(22432,domain/range violations) (11022,undefined predicates) (0,unknown entity types)};
\end{axis}
\end{tikzpicture}
\caption{Conformance debt by kind. Entity typing is clean because its
vocabulary is closed and enumerable; the relationship layer is not, because
relation phrasing is open-ended and every new phrasing is a new predicate. The
domain/range bar is the sum of two separately reported faults, 10{,}025 domain
and 12{,}407 range violations (Table~\ref{tab:debt-counts}). Figures from a
single validator run (\S\ref{sec:limits}).}
\label{fig:debt}
\end{figure} Collapsing
such families back onto a canonical relation is a standing problem in open
knowledge base construction
\cite{galarraga2014canonicalizing, vashishth2018cesi, suchanek2011paris}, and
\S\ref{sec:agent} describes how we intend to use it.

\subsection{When the roles are more correct than the primary}

While reviewing multi-class entities we found a case that argues for keeping
them even when the primary resolver is imperfect.

Our domain extension declares \texttt{StatutoryRule} as a subclass of CCO's
\texttt{Act}, and \texttt{Act} sits under \emph{process}, which is an occurrent.
Statutory rules are consequently classified as things that happen rather than
things that exist: exactly the kind of subsumption link that a discipline
such as OntoClean, or an untangling of the hierarchy into disjoint primitive
branches, is meant to catch
\cite{guarino2002ontoclean, rector2003modularisation}. Government resolutions (documents) inherit that
classification, and forty of the ninety-four multi-class entities we inspected
carried it as their primary type. Those ninety-four are the entities holding two
or more classes at the time of that inspection, taken in full rather than
sampled: a pass earlier than either figure in \S\ref{sec:asymmetry}, which is
why the count matches neither the $3{,}807$ before the evidence rule nor the
$989$ after it. Forty of ninety-four is therefore a census of a small early
population, not an estimate of prevalence in the graph as it now stands.

Their roles told a different story. \emph{Government Resolution} and
\emph{Legislation}, both information content entities, had been attached as
secondary classifications. On these entities the roles were more accurate than
the primary type: the secondary classifications were quietly compensating for a
modelling error one level up.

We have not corrected the parentage, and the reason is worth stating. The
ontology is shared, it is a client deliverable, and re-parenting a class that
forty thousand documents resolve through is a modelling decision with an owner
who is not us: and one whose downstream effects, in a deployed hierarchy, are
not local \cite{rector2011foot, ceusters2006realism}. What the episode
establishes for this paper is narrower: a
representation that records only the winning class would have recorded only the
wrong answer, with nothing left to indicate that a better one had been
available.

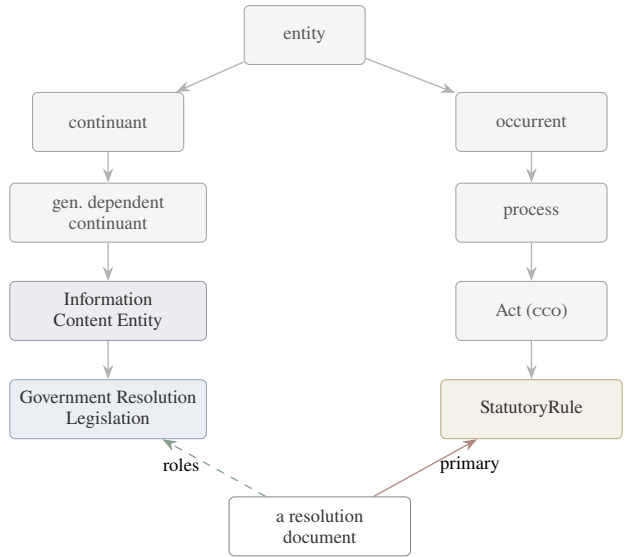
\begin{figure}[H]
\centering
\fitwidth{%
\begin{tikzpicture}[every node/.style={font=\scriptsize}, level distance=0.95cm]
 \node[gdistinct, minimum width=1.6cm] (ent) at (5.0,0) {entity};
 \node[gdistinct, minimum width=2.0cm] (cont) at (2.4,-1.15) {continuant};
 \node[gdistinct, minimum width=2.0cm] (occ) at (8.0,-1.15) {occurrent};
 \node[gdistinct, minimum width=2.6cm] (gdc) at (2.4,-2.35) {gen.\ dependent\\continuant};
 \node[gdistinct, minimum width=2.0cm] (proc) at (8.0,-2.35) {process};
 \node[gonto,   minimum width=2.6cm] (ice) at (2.4,-3.65) {Information\\Content Entity};
 \node[gdistinct, minimum width=2.0cm] (act) at (8.0,-3.65) {Act \textsc{(cco)}};
 \node[groute,  minimum width=2.6cm] (doc) at (2.4,-4.95) {Government Resolution\\Legislation};
 \node[greview,  minimum width=2.4cm] (sr)  at (8.0,-4.95) {StatutoryRule};

 \foreach \a/\b in {ent/cont, ent/occ, cont/gdc, occ/proc, gdc/ice, proc/act, ice/doc, act/sr}
  \draw[gflow, draw=black!30] (\a) -- (\b);

 \node[gdata, minimum width=2.4cm] (e) at (5.2,-6.5) {a resolution\\document};
 \draw[gflow, draw=s5col!75] (e) -- node[right, font=\scriptsize, pos=0.55] {primary} (sr);
 \draw[gflow, draw=s4col!75, dashed] (e) -- node[left, font=\scriptsize, pos=0.55] {roles} (doc);
\end{tikzpicture}}
\caption{The mis-parenting. \texttt{StatutoryRule} descends from CCO's
\texttt{Act} and therefore from \emph{process}, so a document resolves to the
occurrent branch as its primary type while its roles sit correctly under
Information Content Entity. The secondary classifications are the more accurate
ones.}
\label{fig:bfo}
\end{figure}

% ══════════════════════════════════════════════════════════════
\section{Human-gated curation}
\label{sec:human}

\subsection{Flag-only entity resolution}
\label{sec:flagonly}

After the incident of \S\ref{sec:incident} we changed the policy rather than the
threshold. Entity resolution still runs, still computes similarity, and still
ranks candidate pairs \cite{papadakis2020blocking}. It no longer merges anything.
Candidates are written to
an administrative queue and a person decides
\cite{stonebraker2013tamer, wang2012crowder}. At the point of measurement that
queue held 3{,}304 candidate pairs (1{,}002 ranked high-confidence and 2{,}302
medium) none of them applied, and none of them adjudicated either. The $775$
decisions of \S\ref{sec:queue} belong to the ontology-proposal queue; this
queue has had none. We should say plainly what that means, since
\S\ref{sec:queue} levels the same criticism at the other queue: an unworked
queue is a record, not a workflow. This one has never been worked, and until it
is, flagging defers a decision rather than making one.

We should be exact about what that means, because it is not costless and
\S\ref{sec:queue} says as much about the other queue. Flag-only resolution does
not make duplicates go away; it declines to remove them automatically. Up to
$1{,}002$ high-confidence duplicate pairs are therefore live in the graph now.
They inflate the entity count of Table~\ref{tab:coverage}, and they split
retrieval: a question about an entity represented twice reaches whichever copy
the query matched. That is the price of the policy, and we prefer it to the
alternative in \S\ref{sec:incident}: but it is a price, not a clean result,
and an unreviewed dedup queue is no more a safety mechanism than an unreviewed
proposal queue.

This costs us throughput and we accept the trade deliberately. The rule we now
apply when deciding whether to automate a step is not "how accurate is the
classifier" but "what does a wrong answer cost, and would we notice". Tagging a
node with the wrong class is cheap and visible: it shows up in a query, and
re-tagging fixes it. Merging two nodes is expensive and invisible. Those belong
on opposite sides of the automation line regardless of which one the classifier
is better at.

\subsection{Auditability and reversal}

Every manual classification writes an audit record carrying the before and after
state, the editor, and an identifier, and every one can be reversed from it
\cite{buneman2001provenance, lebo2013provo}.
Manually classified nodes are flagged so that a subsequent bulk pass does not
silently overwrite a human decision: a small detail that matters more than it
sounds, because the bulk passes are long-running and the humans are not
watching them.

The asymmetry is worth admitting. Manual classification is audited and
reversible; automated bulk tagging is neither. A tagging run writes classes to
half a million entities and leaves no record of what it changed, so the only way
to undo one is to run another. That sits oddly beside the per-run reversal we
require of the agent in \S\ref{sec:agent}, and the inconsistency is ours rather
than principled: the agent was designed after we had learned to want that
property, and the tagger was never retrofitted with it. Stamping a run
identifier on every write the tagger makes is the obvious repair.

\subsection{The proposal queue and why it stopped working}
\label{sec:queue}

Extraction does not only produce entities. When it encounters a concept the
ontology cannot express, it proposes one: a candidate term with its evidence,
the documents it appeared in, a suggested parent, and the domain and range
observed in text \cite{noy2001ontology101}. Proposals are not applied. They wait
in a review queue.

The mechanism works. What stopped working is the arithmetic around it. Humans
have accepted $722$ proposals and rejected $53$. The queue currently holds
$48{,}403$ pending items ($42{,}820$ relations, $2{,}905$ entities and
$2{,}678$ data properties) against roughly $5{,}900$ when we first looked at
it. Inflow exceeds review by a wide and growing margin (\S\ref{sec:queue}
gives the rates and explains why we do not quote a multiple) and the gap
widens with every document processed. That manual
curation does not scale with automated extraction is not a new observation; it
is the same arithmetic that biocuration ran into two decades ago
\cite{baumgartner2007manual, hirschman2012biocuration}.

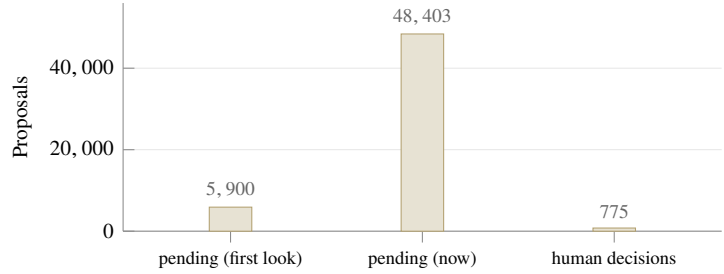
\begin{figure}[H]
\centering
\fitwidth{%
\begin{tikzpicture}
\begin{axis}[
 width=0.68\textwidth, height=4.6cm,
 ybar, bar width=16pt,
 ymin=0, ymax=56000,
 ylabel={Proposals}, ylabel style={font=\footnotesize},
 scaled y ticks=false,
 /pgf/number format/1000 sep={,},
 symbolic x coords={pending (first look), pending (now), human decisions},
 xtick=data,
 xticklabel style={font=\scriptsize}, yticklabel style={font=\footnotesize},
 nodes near coords, nodes near coords style={font=\scriptsize, text=black!60},
 enlarge x limits=0.28,
 axis lines*=left,
 axis line style={draw=black!45, line width=0.4pt},
 tick style={draw=black!45, line width=0.4pt},
 ymajorgrids, grid style={black!10, line width=0.3pt},
]
\addplot+[draw=bugcol!80, fill=bugcol!25] coordinates
 {(pending (first look),5900) (pending (now),48403) (human decisions,775)};
\end{axis}
\end{tikzpicture}}
\caption{The curation queue against human throughput. Total human decisions to
date (acceptances and rejections combined) are shown on the right at the
same scale.}
\label{fig:queue}
\end{figure}

An unreviewed queue is not a safety mechanism. It looks like one, because
nothing is being applied without approval, but a backlog nobody reads is just
signal being discarded slowly. Every unreviewed term is also a compounding loss:
accepted vocabulary feeds back into extraction prompts and into typed retrieval,
so a proposal that sits for a month is a month of documents processed against a
model poorer than it needed to be.

The dynamics are worth stating plainly, because they decide whether curation is
a solved problem or a losing one \cite{little1961proof, kleinrock1975queueing}.
Writing $\lambda$ for proposals arriving per
day and $\mu$ for the rate a reviewer sustains, the backlog evolves as

\begin{equation}
\frac{\mathrm{d}Q}{\mathrm{d}t} = \lambda - \mu,
\qquad
Q(t) = Q_0 + (\lambda - \mu)\,t
\label{eq:queue}
\end{equation}

Here $\lambda \approx 10^{3}$ proposals per day, inferred from queue growth,
and $\mu$ takes two values that must not be confused. There is no equilibrium
when $\lambda > \mu$: the queue does not stabilise at some large value, it
diverges linearly, and every additional corpus raises $\lambda$.

The \emph{observed} review rate is what actually happened: $775$ decisions in
total across the whole period the queue has existed, against inflow of
$43{,}278$ items (net growth plus the decisions themselves) counted from
the first observation at $5{,}900$. The two windows differ: the decisions span
the queue's whole life, the inflow only the period since that first look, which
at $\lambda \approx 10^{3}$ is roughly six weeks. The $1.8\%$ therefore
flatters the review rate, and the share of arrivals reviewed over the matching
window is lower still. Either way it is tens of decisions per day, not
hundreds. The \emph{sustainable} rate $\mu \approx 2\times10^{2}$ per
day is an estimate of what one reviewer could hold if the task were their whole
job. It is an assumption, not a measurement.

We therefore state the qualitative conclusion and decline the arithmetic that
would dress it up. That $\lambda > \mu$, and that the backlog consequently
diverges rather than settling, follows from the observed queue growth alone and
does not depend on either figure being right. The specific multiples we gave in
an earlier draft (five against the assumed rate, fifty against the observed
one) combine an inferred $\lambda$ with an assumed $\mu$ over an unrecorded
window, and we withdraw them. They read as findings and are closer to
illustrations.

Either way the conclusion is the same and the arithmetic is worth stating
plainly. Against the sustainable bound the gap is roughly $5\times$; against
what was actually reviewed it is closer to $50\times$. We report both because
the second is the operating reality and the first is the best case that
additional staffing could reach. The only terms available to change are
$\lambda$, which we do not control, and the fraction of the queue that requires
a human at all.

We note one gap in this analysis: the observation window was not recorded, so
$\lambda$ is inferred from queue growth rather than measured directly, and a
reader cannot check it independently. Instrumenting arrival and decision
timestamps is a small change we should have made at the outset.

% ══════════════════════════════════════════════════════════════
\section{A triage agent for the proposal queue}
\label{sec:agent}

The queue is large but it is not uniformly hard. Sampling it shows three
populations of very different character. A substantial fraction is junk: document numbers, dates and sentence fragments that the extractor mislabelled as
concepts. A larger fraction is near-duplicate: the lending-predicate family
mentioned in \S\ref{sec:conformance} is a dozen queue entries describing one
relation. What remains, and it is a minority, is genuinely new vocabulary that
deserves a considered decision.

Only the third population needs a human. The first two need rules applied
consistently, which is what machines are for
\cite{pan2024unifying, zhu2024llmskg, giglou2023llms4ol}. This section describes
an agent
we have designed to make that split. We are explicit at the outset that it is
designed and scaffolded, not deployed: no proposal in the figures above was
decided by it. We include it because the shape of the design is the
contribution, and because it follows directly from the policy in
\S\ref{sec:flagonly}.

\subsection{Structure}

The agent is a state machine over batches of proposals rather than a
free-running reason-and-act loop \cite{yao2023react}, so that every transition is
inspectable and the whole run is
resumable from a checkpoint after a crash.

\begin{figure}[H]
\centering
\fitwidth{%
\begin{tikzpicture}
 \node[gtiny, text width=2.6cm, align=center] at (4.0,1.35)
    {junk regexes $+$ exact catalog match; no model call};
 \node[gtiny, text width=2.6cm, align=center] at (7.6,1.35)
    {greedy centroid, $s\ge0.90$; expected $\rho=5$--$15$};

 \node[grec] (in) at (0,0) {\begin{tabular}{@{}r@{\hskip 4pt}l@{}}
  42\,820 & relations\\ 2\,905 & entities\\ 2\,678 & data properties\\
 \end{tabular}};
 \node[gproc, minimum width=2.3cm] (gate) at (4.0,0) {mechanical gate};
 \node[gproc, minimum width=2.3cm] (clust) at (7.6,0) {cluster};
 \node[gproc2, minimum width=2.1cm] (judge) at (11.0,0) {judge};

 \node[gdiamond, text width=1.4cm] (val)  at (11.0,-2.8) {validator};
 \node[gmerge,  minimum width=2.0cm] (commit) at (6.6,-2.8) {commit};
 \node[goutput, minimum width=2.0cm] (report) at (2.2,-2.8) {report};

 \node[grec] (checks) at (11.0,-5.2) {\begin{tabular}{@{}l@{}}
  naming convention\\ parent resolves\\ merge target exists\\
  no name collision\\ domain / range real\\
 \end{tabular}};
 \node[gdistinct, minimum width=2.2cm] (human) at (6.6,-5.2) {needs human};

 \draw[gflow] (in)  -- (gate);
 \draw[gflow] (gate) -- node[above, font=\scriptsize, text=black!55] {survivors} (clust);
 \draw[gflow] (clust) -- (judge);
 \draw[gflow] (judge) -- (val);
 \draw[gflow] (val)  -- node[above, font=\scriptsize, text=black!55] {clean} (commit);
 \draw[gflow] (commit) -- (report);
 \draw[gflow] (commit) -- node[right, font=\scriptsize, text=black!55] {twice failed} (human);
 \draw[gflow, draw=black!30] (val) -- (checks);
 \draw[gflow] (gate.south) -- (4.0,-1.9) -| node[near start, above, font=\scriptsize, text=black!55] {junk / dup} (commit.north);
 \draw[gflow, dashed] (val.east) -- (13.1,-2.8) -- (13.1,0)
    node[midway, right, font=\scriptsize, text=black!55] {one repair} -- (judge.east);
\end{tikzpicture}}
\caption{Triage topology. The dashed edge is the batch loop. Two paths reach a
terminal state without a model call at all: mechanically rejected junk, and
exact duplicates of existing vocabulary.}
\label{fig:agent}
\end{figure}
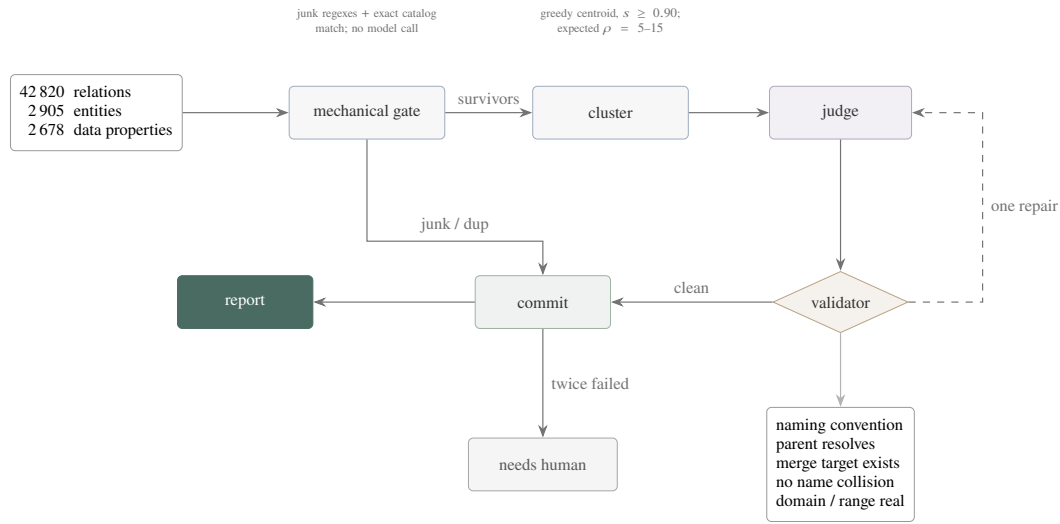

\subsection{Spending model calls only where judgement is required}

The design principle is that inference is the expensive, least predictable step,
so everything decidable without it is decided without it.

The mechanical gate is regular expressions and normalised string matching:
bare numbers, dates, document references, anything implausibly long, and
anything whose normalised form already exists in the vocabulary. None of this
requires a model and all of it is unit-testable.

Survivors are then clustered by embedding \cite{reimers2019sbert} before any
inference happens, which is
the step that makes the economics work: the same compression that predicate
canonicalisation relies on \cite{galarraga2014canonicalizing, vashishth2018cesi}.
Proposals are embedded, greedily
assigned to centroids at cosine similarity $\ge 0.90$, and each cluster is
represented by its most-observed member. If $N$ proposals form $K$ clusters, the
judge sees $K$ decisions rather than $N$:

\begin{equation}
\rho = \frac{N}{K}
\end{equation}

Twelve variants of one lending predicate become a single judgement, and the
eleven others are recorded as synonyms of whatever that judgement decides.

We expect $\rho$ between $5$ and $15$ on the relation-heavy queue. That range is
an extrapolation from the existing predicate-canonicalisation code, which
clusters the same family of terms at the same threshold, and not a measurement
of the agent: the agent has never run. A dry-run pass over the $48{,}403$
pending items would settle it without writing anything, and is the obvious first
experiment.

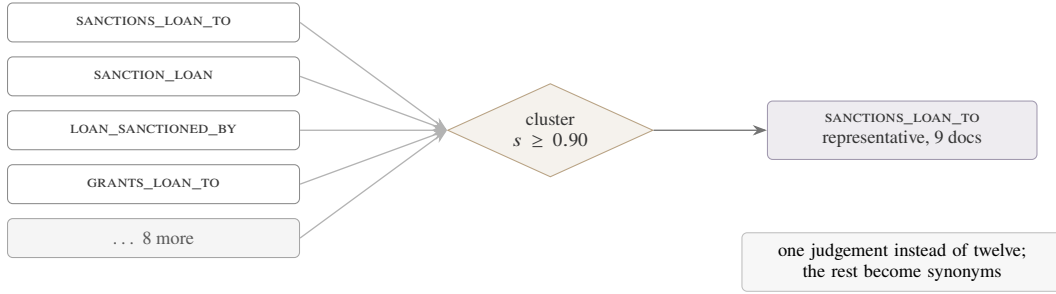
\begin{figure}[H]
\centering
\fitwidth{%
\begin{tikzpicture}[every node/.style={font=\scriptsize}]
 \node[gdata, minimum width=3.9cm, minimum height=0.52cm] (p1) at (0,0)   {\textsc{sanctions\_loan\_to}};
 \node[gdata, minimum width=3.9cm, minimum height=0.52cm] (p2) at (0,-0.72) {\textsc{sanction\_loan}};
 \node[gdata, minimum width=3.9cm, minimum height=0.52cm] (p3) at (0,-1.44) {\textsc{loan\_sanctioned\_by}};
 \node[gdata, minimum width=3.9cm, minimum height=0.52cm] (p4) at (0,-2.16) {\textsc{grants\_loan\_to}};
 \node[gdistinct, minimum width=3.9cm, minimum height=0.52cm] (p5) at (0,-2.88) {\dots\ 8 more};

 \node[gdiamond, text width=1.5cm] (c) at (5.3,-1.44) {cluster\\$s \ge 0.90$};
 \node[gonto, minimum width=3.6cm] (rep) at (10.0,-1.44)
    {\textsc{sanctions\_loan\_to}\\representative, 9 docs};

 \draw[gflow, draw=black!30] (p1.east) -- (c.west);
 \draw[gflow, draw=black!30] (p2.east) -- (c.west);
 \draw[gflow, draw=black!30] (p3.east) -- (c.west);
 \draw[gflow, draw=black!30] (p4.east) -- (c.west);
 \draw[gflow, draw=black!30] (p5.east) -- (c.west);
 \draw[gflow] (c) -- (rep);
 \node[gnote, text width=4.0cm] at (10.0,-3.2)
    {one judgement instead of twelve;\\the rest become synonyms};
\end{tikzpicture}}
\caption{Clustering before inference. Compression on the relation-heavy queue
runs $5$--$15\times$, so the judge sees hundreds of decisions rather than tens
of thousands.}
\label{fig:cluster}
\end{figure}

\subsection{The model proposes; the validator disposes}

The judge sees a cluster, the five most similar terms already in the ontology,
and the domain/range pairs actually observed for it in documents
\cite{zheng2023judge}. It returns a
structured verdict \cite{willard2023guided} (accept, merge into an existing
term, reject, or defer to a
human) with a canonical name, a parent, a definition and a confidence.

Nothing it returns is written directly. Its output passes through a
deterministic validator that checks what a model is bad at
\cite{huang2025hallucination} and a program is
good at: naming convention for the term kind, whether the proposed parent
actually exists, whether a merge target resolves, whether the name collides with
something already present, whether the declared domain and range are real
classes, and whether any datatype is in the permitted set. A verdict that fails
is returned once with the specific errors attached, and the judge gets a single
chance to repair it. A second failure routes the cluster to a human: the same
escalation path that existed before the agent, which is the point.

\begin{algorithm}[H]
\caption{Triage of one proposal batch}
\label{alg:triage}
\begin{algorithmic}
\Require pending batch $B$; ontology $O$; dry-run flag $d$
\Comment{every writing stage takes $d$ and records only when $d$ is false}
\State $B \gets \{\, b \in B: \mathrm{status}(b) = \textsf{pending} \,\}$ \Comment{Table~\ref{tab:policy}, row 1}
\State $S \gets{}$ \Call{Survivors}{$B$, $O$, $d$} \Comment{rows 2--4, no model call}
\State $S \gets{}$ \Call{MergeNearest}{$S$, $O$, $d$} \Comment{row 5: $s \ge 0.95$, no model call}
\State $\mathcal{K} \gets{}$ \Call{Cluster}{$S$} \Comment{cosine $\ge 0.90$; representative by observation count}
\ForAll{$K \in \mathcal{K}$}
 \State $C \gets{}$ \Call{Context}{$K$, $O$} \Comment{5 nearest terms $+$ observed domain/range}
 \State $v \gets{}$ \Call{Judge}{$K$, $C$}
 \State $E \gets{}$ \Call{Validate}{$v$, $O$}
 \If{$E \neq \emptyset$}
  \State $v \gets{}$ \Call{Judge}{$K$, $C$, $E$} \Comment{single repair attempt, errors attached}
  \State $E \gets{}$ \Call{Validate}{$v$, $O$}
 \EndIf
 \If{$E \neq \emptyset$ \textbf{ or not } \Call{PolicyAccepts}{$v$}}
  \State \Call{DeferToHuman}{$K$, $v$, $d$} \Comment{Table~\ref{tab:policy}, row 9}
 \ElsIf{\textbf{not} $d$}
  \State \Call{Commit}{$v$, $O$} \Comment{idempotent; guarded on status = pending}
 \EndIf
\EndFor
\end{algorithmic}
\end{algorithm}

The repair loop is bounded at one iteration by construction rather than by a
counter that could drift: the body contains exactly one re-judge, and the branch
that follows it catches both a still-invalid verdict and a valid verdict the
policy declines to act on.

\begin{table}[H]
\centering
\caption{Verdict policy, evaluated in order. Every threshold is configuration,
not code. $s$ is cosine similarity to the nearest existing term; \emph{seen} is
the number of documents a cluster was observed in.}
\label{tab:policy}
\small
\begin{tabular}{@{}cP{5.0cm}P{4.4cm}@{}}
\toprule
\# & Condition & Outcome \\
\midrule
1 & already decided by a human & leave untouched \\
2 & matches a junk pattern & reject, no model call \\
3 & exact or normalised duplicate & reject as duplicate \\
4 & duplicate, but a new surface form & merge; record as synonym \\
5 & $s \ge 0.95$ & merge, no model call \\
6 & $0.80 \le s < 0.95$ & judge decides merge vs.\ new; \emph{merge} is terminal, \emph{new} continues to row 7 \\
7 & judge accepts, confidence $\ge 0.80$, validator clean, \emph{seen} $\ge 3$ (5 for relations), $s < 0.80$\textsuperscript{$\dagger$} & accept \\
8 & judge rejects with confidence $\ge 0.90$, or \emph{seen} $=1$ & reject \\
9 & anything else & defer to human, with the agent's reasoning attached \\
\bottomrule
\end{tabular}

\vspace{2pt}
{\footnotesize $\dagger$ Rows 5 and 6 have already consumed everything at
$s \ge 0.80$, so this guard is implied; it is written out so the row reads
standalone.}
\end{table}

Row 9 is the one we care about most. The agent's usefulness is not that it
decides everything; it is that whatever it cannot decide arrives at a human
already clustered, already contextualised, and annotated with why it was
punted.

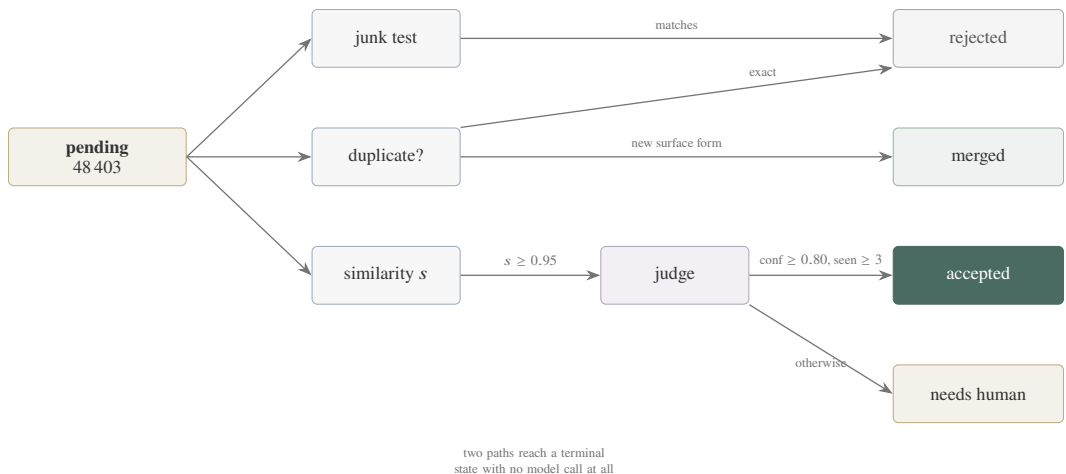
\begin{figure}[H]
\centering
\fitwidth{%
\begin{tikzpicture}
 \node[greview, minimum width=2.4cm] (pend) at (0,-1.9) {\textbf{pending}\\48\,403};

 \node[gproc, minimum width=2.0cm] (gate) at (3.9,-0.3) {junk test};
 \node[gproc, minimum width=2.0cm] (dup) at (3.9,-1.9) {duplicate?};
 \node[gproc, minimum width=2.0cm] (sim) at (3.9,-3.5) {similarity $s$};
 \node[gproc2, minimum width=2.0cm] (judge) at (7.8,-3.5) {judge};

 \node[gdistinct, minimum width=2.3cm] (rej) at (11.9,-0.3) {rejected};
 \node[gmerge,  minimum width=2.3cm] (mer) at (11.9,-1.9) {merged};
 \node[goutput,  minimum width=2.3cm] (acc) at (11.9,-3.5) {accepted};
 \node[greview,  minimum width=2.3cm] (hum) at (11.9,-5.1) {needs human};

 \draw[gflow] (pend.east) -- (gate.west);
 \draw[gflow] (pend.east) -- (dup.west);
 \draw[gflow] (pend.east) -- (sim.west);

 \draw[gflow] (gate.east) -- node[above, font=\tiny, text=black!55] {matches} (rej.west);
 \draw[gflow] (dup.east) -- node[above, font=\tiny, text=black!55] {new surface form} (mer.west);
 \draw[gflow] (dup.north east) -- node[above, font=\tiny, text=black!55, pos=0.7] {exact} (rej.south west);
 \draw[gflow] (sim.east) -- node[above, font=\tiny, text=black!55] {$s\ge0.95$} (judge.west);
 \draw[gflow] (judge.east) -- node[above, font=\tiny, text=black!55] {conf $\ge0.80$, seen $\ge3$} (acc.west);
 \draw[gflow] (judge.south east) -- node[below, font=\tiny, text=black!55] {otherwise} (hum.west);

 \node[gtiny, text width=3.4cm, align=center] at (5.9,-6.0)
    {two paths reach a terminal state with no model call at all};
\end{tikzpicture}}
\caption{The verdict policy of Table~\ref{tab:policy} as a state machine. A
proposal leaves \textsf{pending} only through a guarded transition, and the
guards are ordered so that the cheapest decisive test runs first.}
\label{fig:lifecycle}
\end{figure}

\subsection{Earning the right to write}

The agent runs read-only by default. Producing a report is the normal mode;
writing requires an explicit flag, and the commit path asserts it rather than
trusting the caller.

Before writes are enabled at all, a run's verdicts are sampled and checked by a
person, and we require agreement on at least $28$ of $30$ sampled decisions with
no incorrect acceptances. That fraction is a point estimate near $93\%$ with a
lower $95\%$ bound around $78\%$; for a component writing to a shared client
deliverable the bound is the number that matters, and thirty samples is a weak
gate we would widen before trusting it broadly. Terms the agent creates are marked with its identity
and a run identifier, so a run can be reversed as a unit: created terms removed,
proposal states restored, added synonyms stripped. Every write is idempotent, so
re-running a batch changes nothing, and every proposal update is guarded on the
proposal still being pending, so a human editing the same item concurrently
always wins.

None of this is novel machinery. It is the same posture as
\S\ref{sec:flagonly}, applied to a component that writes to the ontology rather
than to the data graph: automate the mechanical, escalate the judgement, and
make every automated write reversible and attributable.

% ══════════════════════════════════════════════════════════════
\section{Limitations}
\label{sec:limits}

The measurements here come from two document families, and we should have said
so earlier. The graph-scale figures (Table~\ref{tab:coverage}, the
conformance counts, the queue sizes) come from a corpus of Maharashtra
government resolutions. The structured-record material comes from a defence
document set with declared table schemas: the identifier examples of
\S\ref{sec:identity}, the incident of \S\ref{sec:incident}, the positional
anecdote, and the worked resolution of \S\ref{sec:resolve} are all from that
family, as is the separate evaluation corpus of \S\ref{sec:edgeid} and
\S\ref{sec:insertionpath}. The two exercise different parts of the design (tables and declared identifiers on one side, free prose and a large ontology on
the other) and no measurement here spans both. The document type is unusually regular, and we
would expect the name-evidence rule of Equation~\ref{eq:evidence} in particular
to behave differently on prose where entity names are less self-describing.

Authority ordering has a failure mode we have observed and not addressed. A
hallucinated table row once reproduced the coordinates, speed and heading of the
row above it verbatim while carrying its own identifier and name. Because the
ladder stops at the first available key, the identifier settled it and the row
was written as a new record; the positional rung, which would have recognised
that two tracks cannot occupy one point, is unreachable for any row that has an
identifier. A fabricated identifier therefore defeats the ladder by construction.
Consulting lower rungs for contradiction rather than stopping at the first match
would catch it, at the cost of the property that makes the ladder cheap.

The multi-class recall question of \S\ref{sec:asymmetry} is open, and we have
now been wrong about it twice. It is not caused by the rule reaching
type-derived candidates, which the code rules out. Nor is it dissolved by the
difference between assignment and entity counts: our conversion ratio is a
blended mid-pass figure and therefore a lower bound on the pre-rule ratio, so
that correction is worth an unknown amount and the residual is at most about
fifteen points rather than exactly fifteen. We attribute the remainder
provisionally to the sample's non-uniformity. Settling it needs a uniform draw
and the pre- and post-rule class-count distributions, which we did not retain.

The conformance figures in \S\ref{sec:conformance} were captured in a single
validator run against a graph that was being written to at the time, and the
relationship tagging pass in Table~\ref{tab:coverage} was interrupted by an
infrastructure outage rather than completing. Both should be re-measured on a
quiescent graph before they are relied on.

The agent of \S\ref{sec:agent} is a design. It has not processed the queue, and
we make no claim about the verdict distribution it would produce; the thresholds
in Table~\ref{tab:policy} are starting points chosen from the behaviour of the
existing predicate-canonicalisation code, not values learned from agent runs.

\subsection{What is missing, and what it would take}

The largest gap is that the ladder is unevaluated. An identity benchmark would
need a few hundred hand-labelled record pairs drawn from both document families,
scored for precision and recall against the similarity matcher it replaced, and
an ablation showing what each rung contributes. We can say what it would
contain; we have not built it. The evidence-rule result needs the same treatment
on a uniform draw, which is a re-run with a different sampler rather than new
machinery. Both are cheap relative to what they would settle, and their absence
is why this paper should be read as a design report supported by measurements
rather than as an evaluation.

\subsection{Artifact availability}

The Maharashtra government resolutions are public records and a subset can be
released. The ontology extension and the conformance validator are separable
from the deployment and can be released with it. The defence document set cannot
be, so the identity material is not independently reproducible; a synthetic
substitute preserving the table structure would be, and we have not built one.

% ══════════════════════════════════════════════════════════════
\section{Conclusion}

The parts of this system that took the longest to get right were not the
difficult-sounding ones. Class resolution against a five-thousand-class ontology
worked early. What kept failing were the joins: deciding whether two records
were the same thing, deciding whether a name licensed a classification, and
deciding when a batch job was actually finished.

Those three failures share a structure. In each case the code was correct
against its own specification, every individual operation succeeded, and the
result was wrong: a merge that was confident and destructive, a classifier
that read a qualifier as a type assertion, a job that inferred completion from
work attempted rather than counting what remained. None of them raised an error.
We found each by looking at outputs that seemed fine and asking whether they
were.

\subsection*{How this system changed}

The shape of the design is the record of those failures. It began by merging
records automatically on a similarity score, because that is what the problem
looks like from the outside. The incident of \S\ref{sec:incident} ended that,
and resolution became flag-only. Applying the same test to edges produced
Equation~\ref{eq:edgeid}: identity from what a thing is, never from where it was
found. Applying it to classification produced Equation~\ref{eq:evidence}, which
separates an assertion from a coincidence of wording. Applying it to the
curation queue itself produced the agent of \S\ref{sec:agent}, whose whole
purpose is to decide only what is mechanical and hand a human the rest, already
sorted.

None of that was designed up front. Each step is the previous step's failure,
generalised.

The design we ended with is less a technique than a rule for where to draw a
line. Automate what is mechanical; escalate what needs judgement; and refuse to
automate operations that are destructive and hard to detect, however good the
classifier gets. The triage agent of \S\ref{sec:agent} is that rule applied to
curation itself: not a system that replaces the reviewer, but one that ensures
the things reaching the reviewer are the things actually worth their attention.

% ══════════════════════════════════════════════════════════════
\appendix
\section{Field issue log}
\label{app:issues}

Issues that shaped the implementation but did not earn space in the main text,
recorded for practitioners. Those analysed in the body (the merge incident,
the backfill, the tagger's pagination) are omitted here.

\subsection*{Issue 1: heterogeneous arrays reject the whole node}

A list property whose elements are of mixed type is refused at write time, and
the refusal fails the entire node rather than the offending property, so one
malformed confidence array silently cost an entity all of its other properties
too.

\paragraph{Resolution.} Element kinds are inspected before writing and a mixed
list is stringified whole (Equation~1). Booleans are classified before numbers:
in Python \texttt{bool} subclasses \texttt{int}, so the natural test merges the
two kinds and defeats the check: an ordering bug we shipped before finding it.

\subsection*{Issue 2: a completion ratio above one}

The ingest indicator reported $98{,}860/98{,}795$ documents processed. The
manifest is append-only by design and is never pruned, so documents ingested and
later moved out of the spool remained in the numerator while leaving the
denominator.

\paragraph{Resolution.} The numerator counts the intersection of manifest and
spool; the remainder is reported separately as archived. Nothing was wrong with
the ingestion: only with the arithmetic describing it.

\subsection*{Issue 3: edge identity derived from provenance}

Edge identities were hashed from the relation type together with the source
file's identifier, so one relationship stated in two documents became two edges.

\paragraph{Resolution.} Equation~\ref{eq:edgeid}: identity from the resolved
endpoints and the normalised relation type only. On the evaluation corpus this
removed 2{,}407 spurious edges: 22.4\% of the
relationship layer as it stood before the fix (equivalently, that layer had been
28.8\% larger than its corrected size).

\subsection*{Issue 4: concurrency raised against the wrong limit}

The embedding stage was slow, so request concurrency was raised twentyfold. It
did not help and cost stability: the work was bounded by the number of round
trips, not by how many were in flight.

\paragraph{Resolution.} Batch the texts rather than parallelise the calls
(\S\ref{sec:insertionpath}).

\subsection*{Issue 5: a shared class mis-parented}

\texttt{StatutoryRule} descends from a class that is an occurrent, so documents
resolve to the process branch as their primary type
(\S\ref{sec:conformance}).

\paragraph{Status.} Unresolved, deliberately. The ontology is shared and is a
client deliverable; re-parenting a class that tens of thousands of documents
resolve through is a modelling decision whose owner is not us. It is recorded
here because the roles on those entities were more accurate than the primary
type, which is an argument for keeping multi-class assignment even where the
primary resolver is imperfect.

\subsection*{Issue 6: an over-correction we diagnosed and could not confirm}

Requiring name-derived class candidates to be anchored in the name
(Equation~\ref{eq:evidence}) improved precision on the sampled entities while
graph-wide multi-class coverage fell from 3{,}807 entities to 989. We recorded
this as an over-correction and attributed it to the rule being applied to
type-derived candidates.

\paragraph{Status.} Partly resolved, and instructive twice over. The attributed
cause does not exist: the admissibility test is gated on the candidate's origin,
so type-derived candidates never reach it. Our second explanation (that the
two rates were merely in different units) was also too quick, and for a
subtler reason. The conversion ratio of $1.013$
(Figure~\ref{fig:classdist}) comes from a mid-pass snapshot, so it blends
already-processed entities with untouched ones and bounds the pre-rule ratio
from below rather than measuring it. The correction is therefore worth an
unknown amount, and the residual is at most about fifteen points. What remains
is provisionally attributed to a sample that over-represents multi-class
entities twofold.
Retained because both errors are the same error: reaching for an explanation
before checking whether the numbers supported one.

% ══════════════════════════════════════════════════════════════
% REFERENCES
% ══════════════════════════════════════════════════════════════

\end{document}